\documentclass[10pt,twocolumn,letterpaper]{article}

\usepackage{iccv}              
\definecolor{iccvblue}{rgb}{0.21,0.49,0.74}
\usepackage[pagebackref,breaklinks,colorlinks,allcolors=iccvblue]{hyperref}
\usepackage{booktabs}
\usepackage{multirow}
\usepackage{amssymb}
\usepackage{pifont}
\usepackage[accsupp]{axessibility}

\def\paperID{5761} 
\def\confName{ICCV}
\def\confYear{2025}

\title{Rethinking the Embodied Gap in Vision-and-Language Navigation: A Holistic Study of Physical and Visual Disparities}

\author{
  Liuyi Wang$^{1,2}$\thanks{Equal contribution},
  Xinyuan Xia$^{2,3}$\footnotemark[1],
  Hui Zhao$^{2}$,
  Hanqing Wang$^{2}$\thanks{Corresponding author},
  Tai Wang$^{2}$,
  Yilun Chen$^{2}$,
  Chengju Liu$^{1,4}$,\\
  Qijun Chen$^{1,4}$\footnotemark[2],
  Jiangmiao Pang$^{2}$ \\
  $^1$Tongji University, $^2$Shanghai AI Laboratory,  $^3$Shanghai Jiao Tong University, \\
  $^4$State Key Laboratory of Autonomous Intelligent Unmanned Systems
}

\begin{document}
\maketitle
\begin{abstract}
Recent Vision-and-Language Navigation (VLN) advancements are promising, but their idealized assumptions about robot movement and control fail to reflect physically embodied deployment challenges. To bridge this gap, we introduce VLN-PE, a physically realistic VLN platform supporting humanoid, quadruped, and wheeled robots. For the first time, we systematically evaluate several ego-centric VLN methods in physical robotic settings across different technical pipelines, including classification models for single-step discrete action prediction, a diffusion model for dense waypoint prediction, and a train-free, map-based large language model (LLM) integrated with path planning. Our results reveal significant performance degradation due to limited robot observation space, environmental lighting variations, and physical challenges like collisions and falls. This also exposes locomotion constraints for legged robots in complex environments. VLN-PE is highly extensible, allowing seamless integration of new scenes beyond MP3D, thereby enabling more comprehensive VLN evaluation. Despite the weak generalization of current models in physical deployment, VLN-PE provides a new pathway for improving cross-embodiment's overall adaptability. We hope our findings and tools inspire the community to rethink VLN limitations and advance robust, practical VLN models. The code is available at \href{https://crystalsixone.github.io/vln_pe.github.io/}{https://crystalsixone.github.io/vln\_pe.github.io}.
\end{abstract}    
\section{Introduction}
\label{sec:intro}

\begin{figure}[t]
    \centering
    \includegraphics[width=\linewidth]{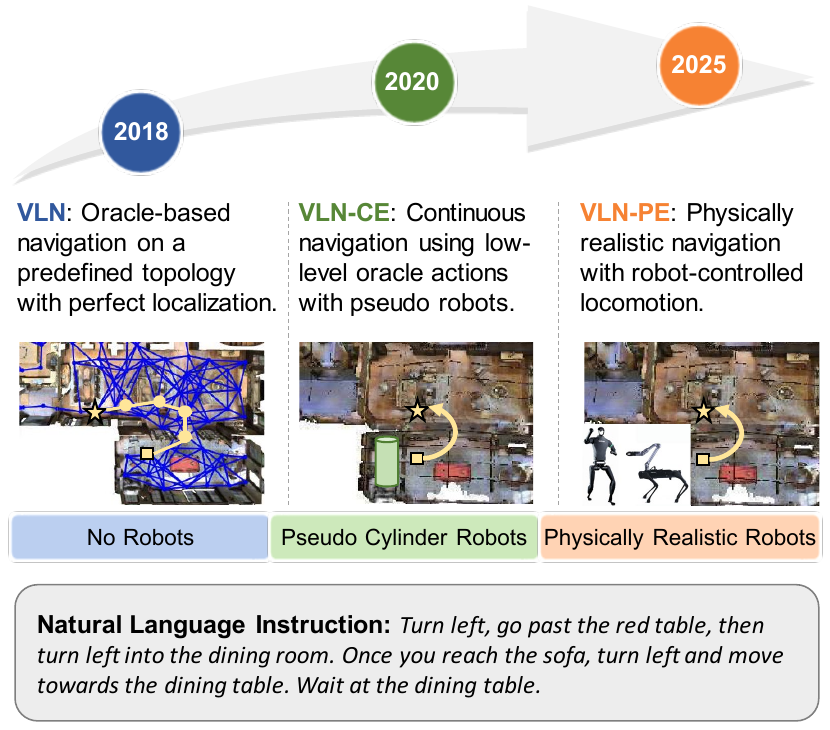}
    \caption{The evolution of vision-and-language navigation (VLN) task. As the techniques and algorithms develop, the settings become more and more practical and challenging.}
    \label{fig:fig1_vlnpr}
\end{figure}

Vision-and-Language Navigation (VLN)~\cite{anderson2018vision} has emerged as a critical task in embodied AI, where agents are required to follow natural language instructions to navigate complex environments. Initially, VLN relied on the MP3D simulator~\cite{anderson2018vision}, which only supported oracle-based navigation by jumping between predefined graph nodes. Later, VLN-CE~\cite{krantz_beyond_2020} introduced continuous navigation using Habitat~\cite{savva2019habitat} (Fig.~\ref{fig:fig1_vlnpr}). A variety of advanced methods ~\cite{zhang2024navid, wang2024causal, an2024etpnav, long2024instructnav} have highlighted the increasing potential of VLN models in advancing the future of embodied AI.

However, a major gap remains between simulation-based models and their physical deployment, especially when applied to diverse robot types, such as wheeled, humanoid, and quadruped robots. Most current VLN-related benchmarks~\cite{ku2020room,qi2020reverie,thomason2020vision,shridhar2020alfred,deitke2022️} are designed for ideal wheeled or point-based agents, ignoring the physical embodiment of the robots themselves. Moreover, the testing conditions in current VLN platforms are overly idealized, often neglecting critical physical issues like viewpoint shifts, falling, deadlocks, and motion errors.
Meanwhile, the rapid advancement of locomotion algorithms, particularly reinforcement learning (RL) methods for humanoid and quadruped robots~\cite{long2024learning,ben2025homie,pan2025roboduet}, has created a growing demand for an integrated VLN benchmarking platform that supports cross-embodiment data collection, training, and evaluation. However, existing research lacks a full-stack platform integrating realistic robotic dynamics, precise locomotion control, and scalable training. Most studies still rely on simplified environments and idealized navigation policies, often based on navigation meshes, which fail to reflect physical complexities. This raises a crucial question: \textit{To what extent do physical embodiment constraints and visual environmental variations impact the performance of existing VLN methods?}

To date, no study has systematically analyzed the applicability of existing VLN methods to different kinds of physical agents. The performance loss remains largely unknown when transferring models from ideal simulated to physical settings. 
To address these gaps, there is a pressing need for a more realistic and adaptable VLN benchmark—one that evaluates language-guided navigation while accounting for the unique locomotion and execution challenges.

\begin{figure*}[t]
    \centering
    \includegraphics[width=\linewidth]{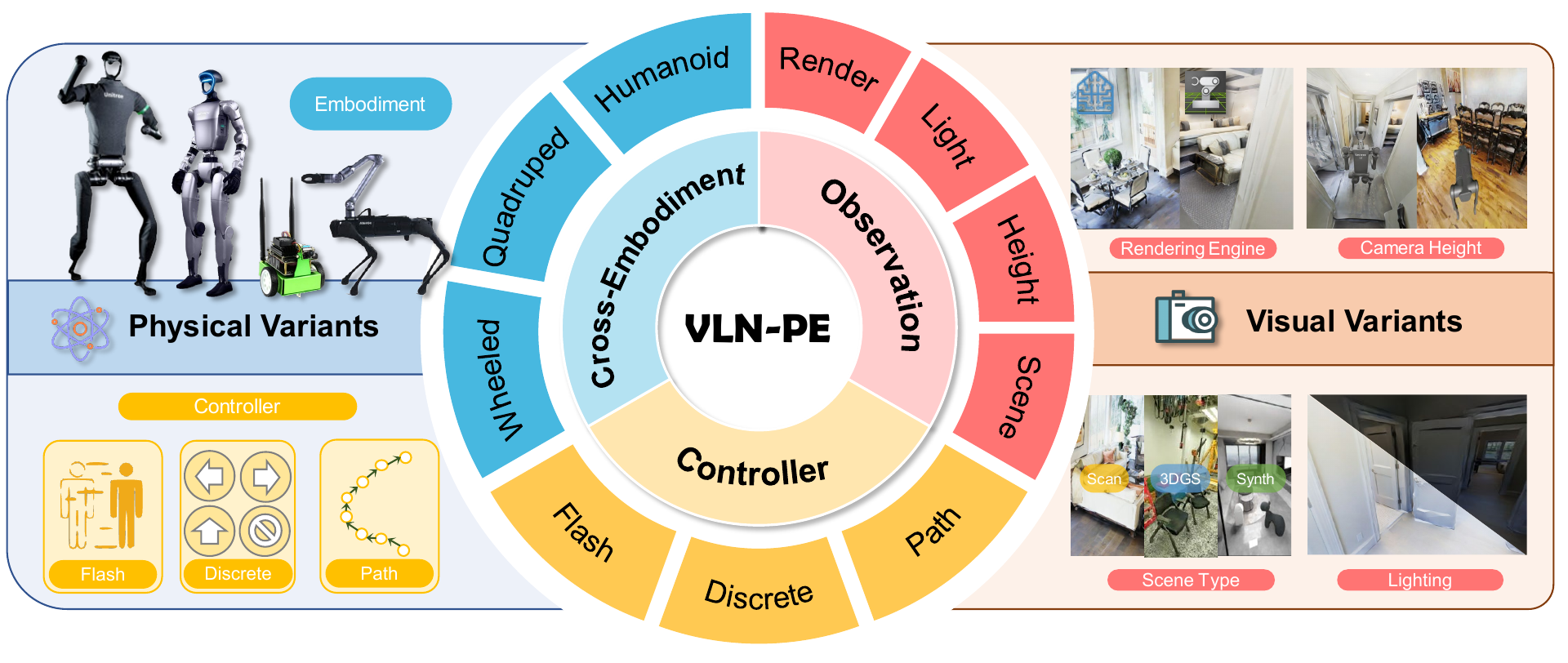}
    \caption{Overview of VLN-PE. This simulator enables researchers to seamlessly integrate various robots and environments, facilitating the exploration of diverse solutions for more physically realistic VLN-related works.}
    \label{fig:overview}
\end{figure*}

In this paper, we introduce VLN-PE (Fig.~\ref{fig:overview}), a physically realistic VLN platform and benchmark that provides a comprehensive environment for cross-embodiment (humanoid, quadruped, and wheeled) data collection and systematic evaluation of policies across various robot embodiments and environmental conditions.
Built on GRUTopia~\cite{wang2024grutopia}, VLN-PE can seamlessly integrate additional environments, including high-quality synthetic scenes and 3D Gaussian Splatting (3DGS) rendering scenes~\cite{xu2023grid,kerbl20233d}, making it highly extensible and user-friendly.
Beyond the widely used MP3D scenes~\cite{chang2017matterport3d}, we introduce several high-quality 3D synthetic household scenes and 3DGS-scanned environments for expanding the scope for VLN research and evaluation.
Specifically, we reveal the influence of the following physical and visual factors on the VLN models:
(1) \textit{Cross-Embodiment Perception}: The effects of robots perceiving and interpreting environments through their unique sensory systems.
(2) \textit{Controller Engagement}: The impact of whether or not to add a physical controller to the data acquisition and validation on model performance.
(3) \textit{Environmental Conditions}: The impact of diverse environmental conditions, such as light intensity and classification.

To align with the typical robot perception setup—usually consisting of a single ego-centric camera—we assess the following ego-centric VLN methods:
(1) Single-step end-to-end methods: Two smaller models, each with approximately 36M parameters—\texttt{Seq2Seq}~\cite{krantz_beyond_2020} and \texttt{CMA}~\cite{krantz_beyond_2020}—along with a fine-tuned, large video-based model, \texttt{NaVid}~\cite{zhang2024navid}, which has 7B parameters.
(2) Multi-step end-to-end method: Diffusion Policy~\cite{chi2023diffusion, cai2025navdp} has shown promise in manipulation tasks but remains underexplored in VLN. We introduce \texttt{RDP}, the first diffusion-based attempt in VLN, as a new baseline method, capable of generating dense trajectory waypoints.
(3) Map-based zero-shot large language model (LLM): \texttt{VLMaps}~\cite{huang23vlmaps}, a zero-shot approach using the LLM to ground the target on a semantic map and navigate using path planning.

Our experiments on VLN-PE reveal several critical insights that highlight limitations in current approaches and suggest promising directions for improvement:
\begin{itemize}
    \item \textit{SoTA Models Struggle in Physical Environments}: Existing VLN-CE models exhibit a 34\% SR relative drop when transferred to physical settings, revealing a gap between pseudo-motion training and physical deployment.
    \item \textit{Cross-embodiment Sensitivity}: Model performance varies across different robots, primarily due to viewpoint height differences, highlighting the need for height-adaptive or perspective-invariant representations.
    \item \textit{Multi-Modal Robustness}: RGB-only models degrade significantly in low-light conditions, whereas RGB + depth models perform more reliably, underscoring the value of multi-modal fusion to improve the model's robustness.
    \item \textit{Limited Generalization of Standard Datasets}: MP3D-style datasets cannot fully capture environment shifts. A simple baseline with 6M trainable parameters, fine-tuned on our small-scale dataset of the newly introduced scenes, outperforms previous SoTA method in zero-shot settings, suggesting the importance of more diverse training distributions and comprehensive evaluation system.
    \item \textit{Towards Cross-Embodiment VLN}: In our experiments, co-training across different robots enables a single baseline to generalize across embodiments and achieve the SoTA result, showing an important foundation for the future unified cross-embodiment VLN model.
\end{itemize}

\section{Related Work}
\label{sec: related_work}

{\bf Vision-and-Language Navigation.}
\label{subsec_vln}
In discrete settings, agents navigate by teleporting between predefined nodes on a navigation graph, which emphasizes cross-modal alignment~\cite{li2023improving,wang2024causal,he2023learning}, semantic understanding~\cite{moudgil2021soat,wang2023dual,cui2023grounded}, and historical dependency~\cite{qiao2023hop_plus,wang2023gridmm,an2024etpnav}.
In contrast, continuous settings (VLN-CE)~\cite{krantz_beyond_2020}, implemented through simulators like Habitat~\cite{ramakrishnan2021habitat}, allow free movement without navigation graphs. While panorama-based methods have shown strong performance in VLN-CE benchmarks~\cite{anderson2021sim,an2024etpnav,hong2022bridging,krantz2022sim}, they heavily rely on panoramic RGB and depth sensors. Since most real-world robots are equipped with egocentric RGB-D pinhole cameras, we focus on solutions that align with this hardware configuration. Traditional approaches~\cite{krantz_beyond_2020, he2024multimodal, wang2024sim,dang2023multiple} focused on training end-to-end specific models, but often struggled with generalization due to simulator-specific constraints. Recent advances have introduced several promising directions: (1) Multimodal Large Language Models (MLLM)~\cite{zhou2024navgpt, kim2024openvla, zhang2024navid, liang2024cornav,cheng2024navila} have shown remarkable potential in cross-modal understanding and generalization; (2) diffusion policies~\cite{chi2023diffusion,sridhar2024nomad,liang2024dtg} have emerged as an effective approach for continuous waypoint prediction in robot manipulation; and (3) map-based methods~\cite{yokoyama2024vlfm, huang2023visual, long2024instructnav} allows zero-shot LLM to incorporate semantic understanding and action planning.
Despite these advances, the challenge of physically realistic cross-embodiment adaptation has received limited attention until recently. Our work reveals this gap through the VLN-PE benchmark, which provides the first comprehensive analysis of how different physical factors impact these methods, offering valuable insights to guide future VLN research directions.

{\bf Language-driven Cross-embodiment Navigation Benchmarks.} GRUTopia~\cite{wang2024grutopia}, built on NVIDIA Isaac Sim, introduces a social navigation task that enables interactions between agents and human-like NPCs. Behavior-100~\cite{srivastava2022behavior} and Behavior-1K~\cite{li2024behavior} focus on various everyday mobile manipulation tasks. 
Yang \textit{et al.}~\cite{yang2024pushing} present a cross-embodiment learning for both manipulation and navigation. ARIO~\cite{wang2024all} standardizes datasets with a unified format, rich sensory modalities, and a mix of real and simulated data, enabling large-scale navigation and manipulation across diverse robots. PARTNR~\cite{chang2024partnr} provides a benchmark for planning tasks in human-robot collaboration, focusing on studying human-robot coordination in household environments. MO-VLN~\cite{liang2023mo} leverages a 3D simulator based on Unreal Engine 5 to evaluate zero-shot multi-task VLN capabilities with wheeled robots. 
While these benchmarks contribute valuable datasets, none systematically evaluate VLN methods across diverse robot embodiments, leveraging the full capabilities of these simulators. In this work, we introduce VLN-PE, a novel simulator that supports multiple robot types for data collection, training, and evaluation. By systematically benchmarking existing ego-centric VLN-CE methods, our study highlights the critical need for more complex and practical VLN settings in future research.
\section{VLN-PE Platform and Benchmark}
\label{sec: data_preparation}

{\bf Simulations.} 
Unlike prior studies that rely on animations and predefined positions for pseudo-action execution, VLN-PE is built on a physically realistic simulator, GRUTopia~\cite{wang2024grutopia}, designed to support various robots. 
It provides RL-based controllers as APIs, enabling the operation of humanoid robots (Unitree H1, G1), quadrupeds (Unitree Aliengo), and wheeled robots (Jetbot). Additionally, thanks to the powerful interactive visualization and rendering capabilities of Isaac Sim, users can easily observe the robot's movement from custom perspectives.

{\bf Scenes.} We converted 90 Matterport3D scenes into USD format. During testing, we identified floor gaps caused by reconstruction errors, which could impact robot movement, particularly for legged robots, whose legs might fall into or get stuck in these holes. To address this, we manually fixed all the holes with the help of volunteers. Next, we aligned the coordinates from the VLN-CE (Habitat) benchmark~\cite{raychaudhuri2021language} to our platform, ensuring that the initial position and rotation matched the original annotations. We employed disk lighting to enhance lighting conditions, allowing us to adjust light intensity for broader research purposes.

Additionally, since MP3D scenes are reconstructed with limited visual diversity, we introduced two additional scene types to further enhance the experimental environment. First, we incorporated 10 high-quality synthetic home scenes from GRScenes~\cite{wang2024grutopia}, which offer significantly improved visual fidelity and physical realism. Second, we included re-built scenes generated using advanced 3D Gaussian techniques~\cite{scaffoldgs}. We use the scanned scene of one laboratory environment, which can be observed via online rendering, providing a highly realistic perceptual experience. 
In theory, many other existing open-sourced scenes can be easily imported into this platform for experimentation.

{\bf Datasets.} The R2R dataset~\cite{anderson2018vision} is the most widely used benchmark for VLN tasks. This dataset is divided into training, validation-seen (same buildings as in training, but with different instructions), and validation-unseen (different buildings from training) splits. We focus on identifying performance gaps using this dataset. 
Currently, our RL-based locomotion controller does not reliably handle stair navigation in complex environments, so we filter out episodes that include stairs. After stair filtering, the \textit{training}/\textit{val-seen}/\textit{val-unseen} splits of R2R remain 8,679/658/1,347 episodes, respectively. 
For the newly introduced scenes, we sampled trajectories and generated VLN-style instructions using a modular LLM method~\cite{he2025navcomposercomposinglanguageinstructions}. After rigorous manual filtering and validation, we created two VLN evaluation datasets: \textit{GRU-VLN10}: 3 scenes for training, 7 for unseen tests, with 441/111/1,287 episodes for training, val-seen, and val-unseen, respectively. \textit{3DGS-Lab-VLN}: A supplementary dataset with 160 training episodes and 640 for evaluation in a 3DGS online rendering environment.

{\bf Metrics.} Following standard VLN evaluation protocols~\cite{anderson2018vision, krantz_beyond_2020}, we use five primary metrics: \textit{Trajectory Length (TL)}, measured in meters; \textit{Navigation Error (NE)}, which quantifies the distance between the predicted and actual stop locations; \textit{Success Rate (SR)}, indicating how often the predicted stop location falls within a predefined distance of the true location; \textit{Oracle Success Rate (OS)}, which assesses the frequency with which any point along the predicted path is within a certain distance of the goal; and \textit{Success Rate weighted by Inverse Path Length (SPL)}, which balances success rate with path efficiency. 
As physical realism is a key focus of this work, we introduce two more metrics: \textit{Fall Rate (FR)}, which measures the frequency of unintended falls, and \textit{Stuck Rate (StR)}, which quantifies instances where the agent becomes immobilized. 

{\bf Locomotion Policy.} 
Our used controller APIs are built upon state-of-the-art RL-based locomotion policies~\cite{long2024hybrid,pan2025roboduet,long2024learning}, specifically designed and optimized for various robot embodiments.
Importantly, the same locomotion models used in simulation can be directly applied in the real-robot experiments, ensuring consistency between virtual and physical deployments. This feature enables more practical and scalable experimentation, significantly reducing both time and cost. Additionally, as our codebase is open-source, researchers are encouraged to explore and integrate more advanced locomotion policies within our framework.
\section{Baselines}
\label{sec: baselines}
Given that most current robots are equipped with ego-centric RGB-D pinhole cameras, the methods we replicate are designed to align with this constraint. We selected three distinct pipelines for replication and performance comparison: (1) end-to-end training methods, including single-step prediction models (CMA~\cite{krantz_beyond_2020}, Seq2Seq~\cite{krantz_beyond_2020}, and NaVid~\cite{zhang2024navid}); (2) multi-step prediction models, with our newly proposed baseline—RDP, based on diffusion policy~\cite{chi2023diffusion}; and (3) a map-based large model retrieval approach (VLMaps)~\cite{huang23vlmaps} that does not require training. 

\subsection{End-to-end Train-based Method.}
\subsubsection{Single-step Discrete Action Classification Methods.}

{\bf Sequence-to-Sequence (Seq2Seq).}
The Seq2Seq baseline~\cite{krantz_beyond_2020} is a straightforward implementation composed of three components: the instruction encoder, which uses an LSTM to encode GLoVE~\cite{pennington2014glove} token embeddings \(I = \{w_i\}_{i=1}^{L}\); the observation encoder, which employs ResNet50~\cite{he2016deep} pre-trained for RGB embedding \(V_t\) and pre-trained on point-goal navigation for depth \(D_t\), respectively; and a recurrent GRU unit takes the combination of the linear maps of these three inputs and predict the next action \(a_t\). The model can be expressed as follows:
\begin{align}
h_t&=\text{GRU}([V_t, D_t, I], h_{t-1}), \\
a_t&=\mathop{\arg\min}\limits_{a} \text{softmax}(W_a h_t + b_a).
\end{align}

{\bf Cross-modal Attention (CMA).}
Built based on Seq2Seq, this method incorporates two recurrent networks - one tracking visual observations as Seq2Seq, and the other making decisions based on attended instructions and visual features. Specifically, the first GRU unit works as $h_t^{1st}=\text{GRU}([V_t, D_t, a_{t-1}],h_{t-1}^{1st})$, where $a_{t-1}\in \mathbb{R}^{1\times 32}$ presents the previous action. Then the instruction features $\hat{I}_t$ attended by the first GRU features are then used to attend to visual ($\hat{V}_t$) and depth ($\hat{D}_t$) features using a scaled dot-product attention mechanism:
\begin{align}
    \hat{I}_t&=\text{Attn}(I, h_t^{1st}),\\
    \hat{V}_T&=\text{Attn}(V_t,h_t^{1st}),\quad \hat{D}_T=\text{Attn}(D_t,h_t^{1st}).
\end{align}
Finally, the second GRU is taken for using the concatenation of these features as inputs and predicting the action:
\begin{align}
    h_t^{2nd} &= \text{GRU}([\hat{I}_t, \hat{V}_t, \hat{D}_t, a_{t-1}, h_t^{1st}], h_{t-1}^{2nd}), \\
    a_t&=\mathop{\arg\min}\limits_{a} \text{softmax}(W_a h_t + b_a).
\end{align}

{\bf Video-based Large Vision-language Navigation Model (NaVid).} 
Compared with the previous two small specific models, NaVid~\cite{zhang2024navid} proposes the first video-based MLLM for VLN in continuous environments, achieving RGB-only navigation akin to human navigation behavior. Specifically, Navid is built based on a general-purpose video-based MLLM named LLaMa-VID~\cite{li2024llama}, consisting of a vision encoder, a query generator, an LLM, and two cross-modality projectors. Given the observations up to time $t$, \textit{i.e.}, a video sequence comprising $t$ frames, NaVid encodes this video to a sequence of tokens via the vision encoder~\cite{sun2023eva} and projects them to a space aligned with language tokens. The special tokens $\texttt{<HIS>}$, $\texttt{<OBS>}$, and $\texttt{<NAV>}$ are adopted to demarcate historical, current observations, and the beginning of the LLM to output the actions in linguistic form.

NaVid's output consists of two variables: an action type, selected from a discrete set, and quantitative arguments for each action. For \textit{Forward}, it predicts the move distance, while for \textit{turn left} and \textit{turn right}, it estimates the rotation degrees. A regular expression parser is used to extract action types and arguments for evaluation and deployment.

\begin{figure}[b]
    \centering
    \includegraphics[width=\linewidth]{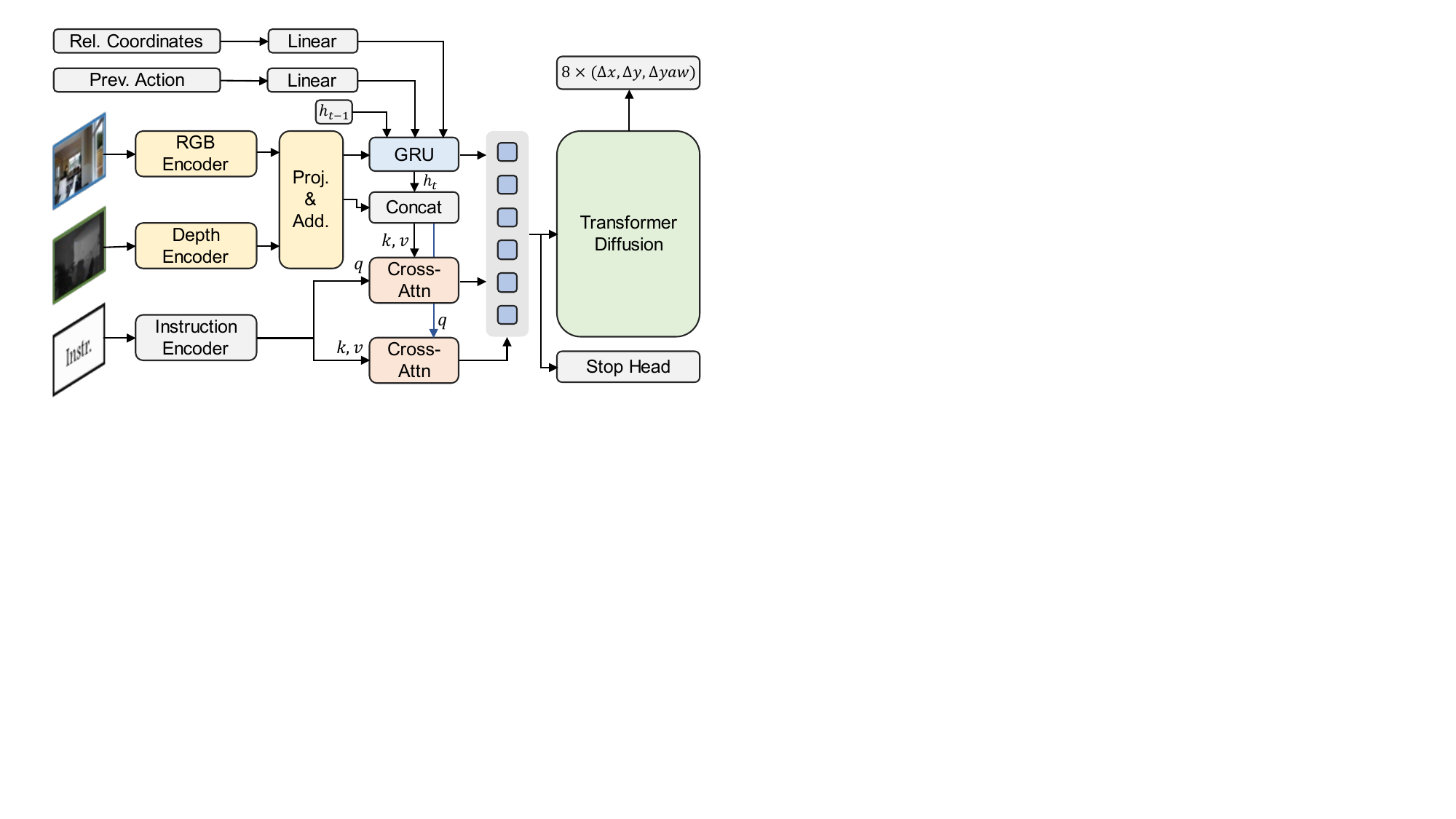}
    \caption{Framework of the recurrent diffusion policy (RDP).}
    \label{fig:RDP}
\end{figure}

\subsubsection{Multistep Continuous Prediction Method.}
{\bf Recurrent Diffusion Policy for VLN (RDP)}.
Recently, diffusion policy~\cite{chi2023diffusion} has demonstrated impressive trajectory smoothness and stability for manipulation tasks. Later, these strategies were applied to point-goal or image-goal navigation tasks~\cite{sridhar2024nomad, bar2024navigation,cai2025navdp}, which inspired our exploration of their potential in the VLN domain. This paper explores \texttt{RDP} (Fig.~\ref{fig:RDP}) as a new baseline method, using the diffusion generative head to support multistep continuous prediction. 

The RDP model takes ego-centric RGB-D input, where RGB and instructions are encoded separately using the LongCLIP~\cite{zhang2024long}, and depth is encoded using a pre-trained ResNet50 model, as in CMA. Information across vision and language is exchanged and aligned using two multi-head, multi-layer cross-modal attention modules~\cite{vaswani2017attention}. The model uses a Transformer structure as the diffusion decoder, taking the fused features $c_t$ as the condition. For the action space, RDP employs continuous relative displacement and yaw angle, represented as a set of continuous ground-truth actions $\{\Delta x_t, \Delta y_t, \Delta{yaw}_t\}_{t=1}^{T}$ for the next $T$ steps.
The overall training and sampling processes follow DDPM~\cite{ho2020denoising}. The iterative denoising process works as follows:
\begin{align}
a_t^{k-1}=\alpha \cdot (a_t^k-\gamma \epsilon_\theta(c_t, a_t^k, k)+\mathcal{N}(0, \mu^2I)),
\end{align} 
where $k$ denotes the number of denoising steps, $\epsilon_\theta$ is a noise prediction network parameterized by $\theta$, and $\alpha$, $\gamma$, and $\mu$ are functions of the noise schedule.

\begin{table*}[t]
\centering
\renewcommand{\arraystretch}{0.85}
\resizebox{\linewidth}{!}{
\begin{tabular}{@{}llcccccccccccccc@{}}
\toprule
\multicolumn{1}{l|}{} & \multicolumn{1}{l|}{} & \multicolumn{7}{c|}{Val Seen} & \multicolumn{7}{c}{Val Unseen} \\
\multicolumn{1}{l|}{\multirow{-2}{*}{Idx}} & \multicolumn{1}{l|}{\multirow{-2}{*}{Method}} & TL$\downarrow$ & NE$\downarrow$ & FR$\downarrow$ & StR$\downarrow$ & OS$\uparrow$ & SR$\uparrow$ & \multicolumn{1}{c|}{SPL$\uparrow$} & TL$\downarrow$ & NE$\downarrow$ & FR$\downarrow$ & StR$\downarrow$ & OS$\uparrow$ & SR$\uparrow$ & SPL$\uparrow$ \\ \midrule
\multicolumn{1}{l|}{1} & \multicolumn{1}{l|}{Random} & 0.14 & 8.24 & 0.30 & 0.00 & 0.30 & 0.30 & \multicolumn{1}{c|}{0.30} & 0.11 & 7.78 & 0.74 & 0.00 & 3.34 & 3.04 & 2.30 \\ \midrule
\multicolumn{16}{c}{\textit{Zero-shot transfer evaluation from VLN-CE}} \\ \midrule
\multicolumn{1}{l|}{2} & \multicolumn{1}{l|}{Seq2Seq-Full~\cite{krantz_beyond_2020}} & 8.29 & 7.59 & 19.15 & 3.04 & 17.63 & 13.83 & \multicolumn{1}{c|}{11.17} & 8.28 & 6.99 & 13.88 & 3.79 & 21.83 & 15.00 & 11.99 \\
\multicolumn{1}{l|}{3} & \multicolumn{1}{l|}{CMA-Full~\cite{krantz_beyond_2020}} & 6.51 & 7.28 & 17.93 & 6.23 & 18.09 & 15.50 & \multicolumn{1}{c|}{14.00} & 6.55 & 7.02 & 15.07 & 4.31 & 19.82 & 16.04 & 14.63 \\
\multicolumn{1}{l|}{4} & \multicolumn{1}{l|}{CMA} & \textbf{4.44} & 7.59 & 21.12 & 1.98 & 13.33 & 11.32 & \multicolumn{1}{c|}{10.20} & \textbf{4.77} & 7.39 & 15.21 & 3.41 & 15.23 & 10.31 & 10.04 \\
\multicolumn{1}{l|}{5} & \multicolumn{1}{l|}{CMA-Height1.8} & 4.87 & 7.71 & 17.17 & 2.28 & 14.06 & 12.87 & \multicolumn{1}{c|}{11.59} & 5.14 & 7.41 & 16.11 & 2.08 & 17.02 & 13.58 & 12.21 \\
\multicolumn{1}{l|}{6} & \multicolumn{1}{l|}{NaVid~\cite{zhang2024navid}} & 7.54 & \textbf{6.20} & \textbf{11.25} & \textbf{0.46} & 24.32 & 21.58 & \multicolumn{1}{c|}{17.45} & 7.12 & \textbf{5.94} & \textbf{8.61} & \textbf{0.45} & 27.32 & 22.42 & 18.58 \\ \midrule
\multicolumn{16}{c}{\textit{Train on VLN-PE}} \\ \midrule
\multicolumn{1}{l|}{7} & \multicolumn{1}{l|}{Seq2Seq} & 9.89 & 7.73 & 22.19 & 3.04 & 30.55 & 19.60 & \multicolumn{1}{c|}{15.67} & 9.51 & 7.91 & 19.67 & 3.71 & 27.62 & 15.89 & 12.58 \\
\multicolumn{1}{l|}{8} & \multicolumn{1}{l|}{CMA} & 10.23 & 7.32 & 23.40 & 2.43 & 31.04 & 21.12 & \multicolumn{1}{c|}{16.15} & 10.09 & 7.43 & 18.63 & 3.12 & 31.33 & 18.78 & 14.56 \\
\multicolumn{1}{l|}{9} & \multicolumn{1}{l|}{RDP} & 12.00 & 7.10 & 22.95 & 3.95 & 33.43 & 23.86 & \multicolumn{1}{c|}{17.35} & 11.76 & 6.97 & 18.75 & 4.58 & \textbf{34.06} & 21.98 & 16.44 \\
\multicolumn{1}{l|}{10} & \multicolumn{1}{l|}{Seq2Seq+} & 9.92 & 7.54 & 26.11 & 5.59 & 31.93 & 19.58 & \multicolumn{1}{c|}{15.13} & 10.21 & 7.64 & 21.82 & 5.12 & 30.47 & 18.13 & 14.06 \\
\multicolumn{1}{l|}{11} & \multicolumn{1}{l|}{\textbf{CMA+}} & 9.09 & 6.68 & 20.21 & 2.74 & \textbf{37.99} & \textbf{28.72} & \multicolumn{1}{c|}{\textbf{24.24}} & 8.61 & 7.11 & 18.63 & 4.83 & 31.55 & \textbf{23.31} & \textbf{19.66} \\ \midrule
\multicolumn{16}{c}{\textit{Train-free Map-based Exploration and Navigation}} \\ \midrule
\multicolumn{1}{l|}{12} & \multicolumn{1}{l|}{VLMaps*~\cite{huang23vlmaps}} & \multicolumn{1}{c}{--} & \multicolumn{1}{c}{--} & \multicolumn{1}{c}{--} & \multicolumn{1}{c}{--} & \multicolumn{1}{c}{--} & \multicolumn{1}{c}{--} & \multicolumn{1}{c|}{--} & \multicolumn{1}{c}{15.73} & \multicolumn{1}{c}{6.98} & \multicolumn{1}{c}{23.00} & \multicolumn{1}{c}{0.00} & \multicolumn{1}{c}{20.00} & \multicolumn{1}{c}{20.00} & 12.70 \\
\bottomrule
\end{tabular}}
\caption{Evaluation of existing ego-centric VLN-CE solutions in VLN-PE using the humanoid robot (Unitree H1) with the RL-based controller for implementation on the R2R dataset~\cite{anderson2018vision}. The units for TL and NE are meters (m), while all other metrics are expressed as percentages (\%). * means that we randomly choose 200 episodes for train-free map-based LLM evaluation.}
\label{tab:performance_of_vlnce}
\end{table*}

Unlike previous applications of diffusion models in robotic manipulation tasks or local navigation tasks, VLN requires more advanced cross-modal understanding, long-term memory, and reasoning abilities. One of the key challenges in designing this model was how to represent historical information. In this paper, we employ a recurrent GRU structure to maintain and update historical observations, enhancing the model's ability to capture long-range dependencies. Additionally, we observe that the diffusion model struggle to determine when to stop based on language inputs. To mitigate this, we introduce an additional MLP prediction head for stop progression, which is continuously updated from 0 to 1 during navigation. This stop branch assists the diffusion model in making more precise stopping decisions. The overall loss function is defined as:
\begin{align}
    \mathcal{L}_{\text{RDP}}=&MSE(\epsilon^k, \epsilon_\theta(c_t, a^0_t+\epsilon^k,k)) \\
    &+ \lambda\cdot MSE(\mathcal{S}_{stop}(c_t), \hat{p}_{stop}),
\end{align}
where $\hat{p}_{stop}$ represents the ground-truth stop progress of each step $\lambda$ is set to 10 through our experiments. For more details, please refer to our Appendix.

\subsection{Modular Map-based Train-free Method.}

{\bf Improved VLMaps.} Train-free, map-based models have demonstrated significant potential, driven by advancements in vision-and-language models with open-set vocabulary capabilities and the versatility of LLMs. A representative work, VLMaps~\cite{huang23vlmaps} leverages LLMs to parse complex natural language commands into a sequence of code-like subgoals that include object and positional relations (\textit{e.g.}, ``robot.move\_in\_between(`sofa', `chair')"). These subgoals are then localized on a semantic map. VLMaps employs LSeg~\cite{lseg} to ground 3D point cloud voxels into semantic embeddings, aligning them with text embeddings of subgoals via contrastive learning. The LLM generates executable Python code, enabling the robot to define functions, logical structures, and parameterized API calls. 
In our experiments, we observed that some landmarks referenced in VLN instructions may not be visible initially. To address this, we incorporate an exploration policy VLFM~\cite{yokoyama2024vlfm} for frontier detection.
Specifically, when the robot fails to detect the target object from the current pixel embeddings, it performs a turn-around maneuver, moves to each frontier, and evaluates whether the viewpoint is likely to contain the next landmark using the image and text encoders from LSeg. 
Please refer to our appendix for more details. 

\section{Experiments}
\label{sec: experiments}
In this section, we aim to answer the following questions:
\begin{itemize}[leftmargin=2em]
\item[(1)] How well do VLN-CE models transfer to VLN-PE?
\item[(2)] How do physical controllers influence performance?
\item[(3)] How does cross-embodiment data impact adaptation?
\item[(4)] How do observation conditions affect accuracy?
\end{itemize}

\subsection{Evaluation of VLN-CE Solutions in VLN-PE}

\begin{figure}
    \centering
    \includegraphics[width=\linewidth]{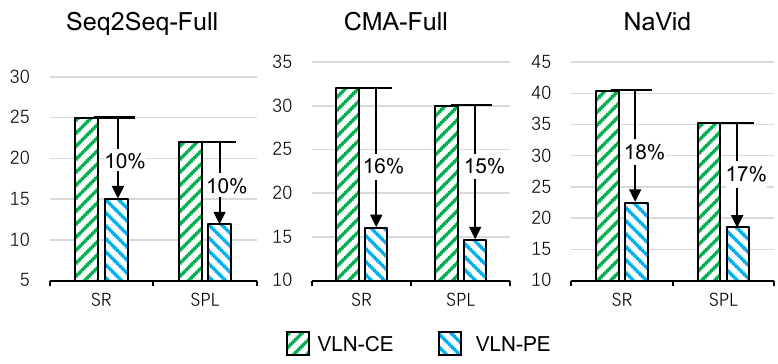}
    \caption{Performance declines in zero-shot testing on the Humanoid H1 robot in VLN-PE on R2R val-unseen.}
    \label{fig:transfer_perf_drop}
\end{figure}

\begin{table*}[t]
\centering
\renewcommand{\arraystretch}{0.85}
\resizebox{\linewidth}{!}{
\begin{tabular}{@{}l|l|ccccccc|ccccccc@{}}
\toprule
\multirow{2}{*}{Idx} & \multirow{2}{*}{Method} & \multicolumn{7}{c|}{Val Seen} & \multicolumn{7}{c}{Val Unseen} \\
 &  & TL$\downarrow$ & NE$\downarrow$ & FR$\downarrow$ & StR$\downarrow$ & OS$\uparrow$ & SR$\uparrow$ & \multicolumn{1}{c|}{SPL$\uparrow$} & TL$\downarrow$ & NE$\downarrow$ & FR$\downarrow$ & StR$\downarrow$ & OS$\uparrow$ & SR$\uparrow$ & SPL$\uparrow$ \\ \midrule
1 & NaVid (ZS) & \textbf{4.88} & 5.05 & 27.02 & \textbf{0.00} & 30.63 & 25.23 & 22.24 & \textbf{5.86} & \textbf{4.75} & \textbf{10.87} & \textbf{0.60} & 23.73 & 18.64 & 13.99 \\ 
2 & CMA-CLIP w/o FT & 5.41 &	5.91 &	68.47 &	6.31 &	18.02& 	10.81 &	9.94 &	6.47 &	5.51 &	55.40 &	12.28 &	22.30 &	15.31 &	13.12 \\ 
3 & CMA-CLIP & 8.95 & \textbf{4.50} & 23.42 & 0.90 & \textbf{57.66} & 31.53 & 27.52 & 9.41 & 5.41 & 27.20 & 5.75 & 46.15 & 22.46 & 17.93 \\
4 & RDP w/o FT & 11.24 & 5.83 & 57.66 & 10.81 & 36.04 & 18.02 & 12.73 & 13.02 & 5.31 & 34.45 & 9.44 & \textbf{51.81} & 26.19 & 18.70 \\
5 & RDP & 10.51 & 4.96 & \textbf{22.52} & 5.41 & 48.65 & \textbf{32.43} & \textbf{27.74} & 9.70 & 4.93 & 13.99 & 5.75 & 45.30 & \textbf{28.52} & \textbf{22.53} 
\\ \bottomrule
\end{tabular}}
\caption{Evaluation of end-to-end methods on the GRU-VLN10 dataset using the Humanoid robot to assess model generalization in an out-of-MP3D-domain setting. \textit{ZS} means the zero-shot implementation. ``w/o FT" means without fine-tuning in the specific scenes.}
\label{tab:gru_vln10}
\end{table*}

\begin{table}[]
\centering
\renewcommand{\arraystretch}{0.85}
\resizebox{\linewidth}{!}{
\begin{tabular}{@{}l|l|ccccccc@{}}
\toprule
\multirow{2}{*}{Idx} & \multirow{2}{*}{Method} & \multicolumn{7}{c}{3DGS-Lab-VLN} \\
 &  & TL$\downarrow$ & NE$\downarrow$ & FR$\downarrow$ & StR$\downarrow$ & OS$\uparrow$ & SR$\uparrow$ & SPL$\uparrow$ \\ \midrule
1 & NaVid (ZS) & 1.44 & 5.70 & \textbf{4.65} & \textbf{0.00} & 6.20 & 5.81 & 1.00 \\
2 & CMA-CLIP w/o FT & 12.06 & 6.40 & 42.75 & 2.54 & 48.78 & 16.72 & 10.66 \\
3 & CMA-CLIP & 8.46 & 5.38 & 11.74 & 0.63 & 30.67 & 24.88 & 17.43 \\
4 & RDP w/o FT & 18.36 & 5.63 & 40.39 & 1.08 & \textbf{53.78} & 26.78 & 12.17 \\
5 & RDP & 9.85 & \textbf{4.73} & 17.97 & 0.63 & 34.84 & \textbf{30.63} & \textbf{22.69} \\ \bottomrule
\end{tabular}}
\caption{Performance on the 3DGS-Lab-VLN val-unseen dataset.}
\label{tab:3dgs_performance}
\end{table}

In Tab.~\ref{tab:performance_of_vlnce}, we evaluate ego-centric VLN-CE solutions using the Humanoid Unitree H1 with an RL-based locomotion controller on the R2R dataset~\cite{anderson2018vision} under three different conditions.
Please note that unless otherwise specified, our experiments are conducted on the R2R dataset.

\textbf{Zero-shot Performance}. As shown in Tab.~\ref{tab:performance_of_vlnce} and Fig.~\ref{fig:transfer_perf_drop}, models transferred directly from VLN-CE to VLN-PE experience significant performance drops. Specifically, Seq2Seq-Full ($\#2$), CMA-Full ($\#3$), and NaVid ($\#6$) see SR declines of 10\%, 16\%, and 18\%, respectively.
Generally, NaVid exhibits better generalization compared to smaller models. As its training code is unavailable, we assess its zero-shot performance as a 7B-parameter model. Despite some performance drop, NaVid achieves the highest zero-shot navigation results, highlighting the potential of MLLMs for VLN.
Additionally, as a zero-shot, map-based LLM solution, VLMaps ($\#12$) demonstrates reasonable results in the VLN-PE setting. This suggests that, beyond end-to-end action prediction, map-based intelligent navigation could also be a viable approach for VLN.

\textbf{In-domain Training and Fine-tuning.} Notably, Seq2Seq-Full and CMA-Full were trained on Habitat with extensive data augmentation, including dagger-based training and EnvDrop-augmented datasets~\cite{tan2019learning}, totaling 175K additional samples. Despite this, they underperform compared to models trained from scratch on in-domain data collected within VLN-PE—without any augmentation ($\#7$ and $\#8$).
Interestingly, fine-tuning Seq2Seq and CMA (denoted as Seq2Seq+ and CMA+) with their SoTA weights using the training dataset from VLN-PE significantly improves performance. In particular, CMA+ ($\#11$) surpasses NaVid's zero shot performance, achieving SR 28.72 \textit{vs.} 21.58 and SPL 24.24 \textit{vs.} 17.45 on val-seen. These results indicate that existing VLN models tend to overfit to specific simulation platforms, resulting in poor generalization when directly transferred to new environments or settings. However, incorporating diverse domain data can further enhance overall navigation performance.
Without additional augmentation, $\#5$ explores the impact of observation height, aligning the agent’s camera height with H1, which improves transferability compared to $\#4$ with the default height of 1.2m. Comparing $\#9$ and $\#7,\#8$, we find that RDP outperforms CMA and Seq2Seq when trained from scratch. This marks the first application of continuous dense low-level offset prediction in VLN, highlighting diffusion policy-based approaches as a promising research direction.

\textbf{Collision Avoidance.} Another notable observation is that the MLLM-based method, NaVid, exhibits significantly lower StR and FR compared to all other methods. This is particularly interesting as it highlights the potential of MLLMs in addressing robotic challenges, such as autonomous obstacle avoidance and deadlock recovery. This advantage can be attributed to the world knowledge capabilities of large models, which allow NaVid to make context-aware decisions about its environment, helping it avoid falls and deadlocks. However, NaVid also has its limitations. We observed that in 70\% of episodes, it tends to rotate continuously near the goal for over 25 steps before issuing a stop signal, rather than proactively stopping when reaching the target. This suggests that large models may still struggle with precise goal recognition, indicating an area for future improvement in MLLM-based navigation.

\textbf{Out-of-MP3D-domain Study.} Since most VLN methods have been developed and evaluated within MP3D environments, we introduce 10 high-quality indoor scenes from GRUScenes~\cite{wang2024grutopia} to expand the evaluation scope. To address the limited vocabulary size in the original CMA model, we replace its instruction encoder with Long-CLIP~\cite{zhang2024long}, denoted as CMA-CLIP. As shown in Tab.~\ref{tab:gru_vln10}, fine-tuning small models ($\#3$ and $\#5$) using just 441 episodes from 3 scene leads to a significant performance boost compared to NaVid’s zero-shot results (\textit{e.g.}, RDP improves SR by 18.86\% and SPL by 17.66\% on average). Surprisingly, on the 3DGS-Lab-VLN dataset (Tab.~\ref{tab:3dgs_performance}), NaVid completely fails, continuously rotating and achieving only 5.81 SR. We suspect this is due to rendering noise introduced by 3DGS, which, while imperceptible to humans, may significantly disrupt the large model’s RGB perception. 
It is important to note that these results do not suggest that small models are sufficient, but rather highlight the substantial room for improvement in current MLLM-based navigation models. 

\subsection{Impact of Physical Controller}
One key advantage of our platform is its support for realistic physical simulation across multiple robot embodiments, particularly legged robots such as humanoid and quadruped robots. 
This raises an important research question: \textit{Can training a VLN model on data collected with physical controller engagement—where robots experience natural walking dynamics and perceptual feedback from motion disturbances—enhance adaptation and performance on legged robots?} To explore this, we conducted comparative experiments using the Humanoid robot in VLN-PE.

\begin{table}[!htb]
\centering
\renewcommand{\arraystretch}{0.85}
\resizebox{\linewidth}{!}{
\begin{tabular}{@{}cc|ccccc|ccccc@{}}
\toprule
\multirow{2}{*}{\begin{tabular}[c]{@{}c@{}}Collect \\ w/ loco\end{tabular}} & \multirow{2}{*}{\begin{tabular}[c]{@{}c@{}}Eval \\ w/ loco\end{tabular}} & \multicolumn{5}{c|}{Val Seen} & \multicolumn{5}{c}{Val Unseen} \\
 &  & FR$\downarrow$ & StR$\downarrow$ & OS$\uparrow$ & SR$\uparrow$ & SPL$\uparrow$ & FR$\downarrow$ & StR$\downarrow$ & OS$\uparrow$ & SR$\uparrow$ & SPL$\uparrow$ \\ \midrule
\ding{55} & \ding{55} & 0.00 & 0.00 & 39.51 & 20.21 & 15.07 & 0.00 & 0.00 & 38.60 & \textbf{21.31} & \textbf{15.89} \\
\checkmark & \ding{55} & 0.00 & 0.00 & \textbf{50.30} & 18.54 & 12.91 & 0.00 & 0.00 & \textbf{43.76} & 15.90 & 10.54 \\
\ding{55} & \checkmark & 31.76 & 4.56 & 18.39 & 12.92 & 11.26 & 25.69 & 4.75 & 19.67 & 13.51 & 11.54 \\
\checkmark & \checkmark & \textbf{23.40} & \textbf{2.43} & 31.04 & \textbf{21.12} & \textbf{16.15} & \textbf{18.63} & \textbf{3.12} & 31.33 & 18.78 & 14.56 \\ \bottomrule
\end{tabular}}
\caption{Comparison of the controller engagement for the data collecting and evaluation in training CMA from scratch.}
\label{tab:controller_effect}
\end{table}

As shown in Tab.~\ref{tab:controller_effect}, the results demonstrate that the model performance is highest when the physical locomotion controller used during data collection and evaluation remains consistent. Conversely, when the controllers differ between training and evaluation, performance drops significantly. Additionally, training with data collected using a controller helps mitigate robot falls and stuck to some degrees, improving overall stability. This demonstrates that for legged robots, the controller-based data collection pipeline provided by our simulator proves to be crucial for achieving more reliable performance.

\subsection{Impact of Cross-embodiment Data}
\begin{figure}[t]
    \centering
    \includegraphics[width=\linewidth]{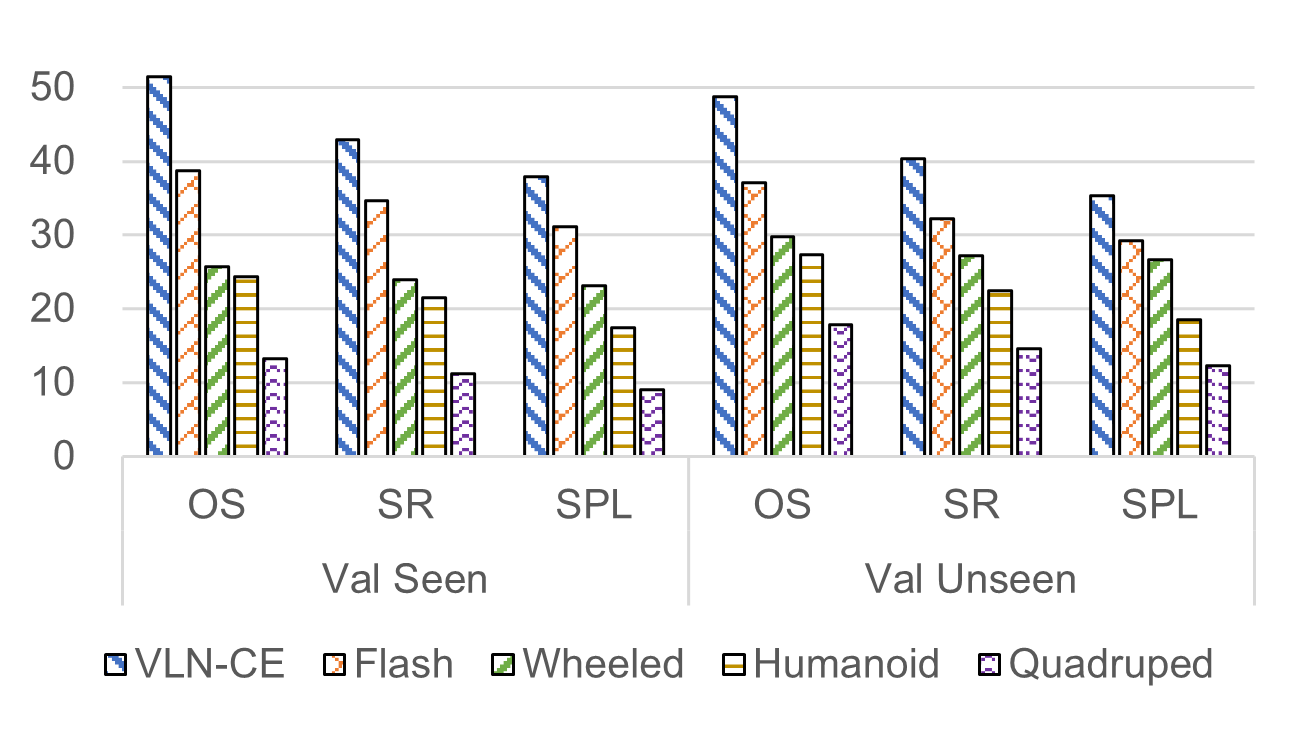}
    \caption{Performance of NaVid across different robot platforms.}
    \label{fig:navid_cross_embodiment}
\end{figure}
Fig.~\ref{fig:navid_cross_embodiment} evaluates NaVid’s zero-shot performance on different robot types, revealing varying degrees of performance degradation. A key advantage of VLN-PE is its ability to seamlessly support diverse robot models for navigation and data collection. This raises our interest: \textit{Can cross-robot data improve model training, enabling a unified model that generalizes across different robot embodiments (``One-for-All")?} To explore this, we collected R2R data from humanoid, quadrupedal, and wheeled robots within VLN-PE. Tab.~\ref{tab:cross_data_training} presents key findings from this study. 

(1) VLN models are highly sensitive to motion dynamics and camera height. When directly transferring models to our platform, the best performance is observed on the Humanoid's flash setting, whose smooth movement is most similar to that in the original Habitat platform. The quadruped robot (about 0.5 m), with a significantly lower camera height, causes the model to nearly fail completely.

(2) Robot-specific viewpoint data enhances model adaptation. Fine-tuning on platform-collected data significantly improves OSR, increasing from 19.82 to 31.33 for humanoid robots and 7.84 to 32.37 for quadrupeds on val-unseen. This underscores the importance of specialized training data for adapting VLN models to different embodiments. Additionally, with limited data, smaller models struggle with accurate stopping—a key challenge in VLN.

(3) Cross-embodiment training significantly enhances overall performance and enables a \textit{One-for-All} model. As indicated by the bolded results, models trained using a combination of data from all three robot types consistently achieve the best performance. This improvement can be attributed both to increased data volume and to the benefits of multi-view learning, which enhances the model's understanding of both the environment and its own embodiment.

\begin{table}[!htb]
    \centering
    \Large
    \renewcommand{\arraystretch}{1}
    \resizebox{\linewidth}{!}{
    \begin{tabular}{@{}l|cccc|ccc|ccc@{}}
\toprule
\multirow{2}{*}{Robot} & \multicolumn{4}{c|}{Training Data} & \multicolumn{3}{c|}{Val Seen} & \multicolumn{3}{c}{Val Unseen} \\
 & \multicolumn{1}{l}{VLN-CE} & \multicolumn{1}{l}{Humanoid} & \multicolumn{1}{l}{Quadruped} & \multicolumn{1}{l|}{Wheeled} & OS & SR & SPL & OS & SR & SPL \\ \midrule
\multirow{4}{*}{Humanoid} & \checkmark &  &  &  & 18.09 & 15.50 & 14.00 & 19.82 & 16.04 & 14.63 \\
 &  & \checkmark &  &  & 31.04 & 21.12 & 16.15 & 31.33 & 18.78 & 14.56 \\
 &  & \checkmark & \checkmark & \checkmark & \textbf{35.26} & 22.64 & 18.07 & 31.70 & 19.30 & 16.97 \\
 & \checkmark & \checkmark & \checkmark & \checkmark & 32.37 & \textbf{26.44} & \textbf{22.76} & \textbf{34.08} & \textbf{26.87} & \textbf{23.54} \\ \midrule
\multirow{4}{*}{Quadruped} & \checkmark &  &  &  & 4.96 & 2.07 & 1.69 & 7.84 & 4.73 & 3.80 \\
 &  &  & \checkmark &  & 31.61 & 17.93 & 14.34 & \textbf{32.37} & 17.00 & 13.40 \\
 &  & \checkmark & \checkmark & \checkmark & \textbf{32.11} & 21.88 & 18.84 & 31.72 & 17.15 & 14.68 \\
 & \checkmark & \checkmark & \checkmark & \checkmark & 30.40 & \textbf{24.47} & \textbf{20.93} & 29.62 & \textbf{23.83} & \textbf{20.75} \\ \midrule
\multirow{4}{*}{Wheeled} & \checkmark &  &  &  & 15.32 & 12.61 & 12.61 & 13.75 & 11.02 & 10.80 \\
 &  &  &  & \checkmark & 16.98 & 15.09 & 14.58 & 14.63 & 12.01 & 11.78 \\
 &  & \checkmark & \checkmark & \checkmark & 21.51 & 19.35 & 18.71 & 21.44 & 15.83 & 14.87 \\
 & \checkmark & \checkmark & \checkmark & \checkmark & \textbf{23.66} & \textbf{20.50} & \textbf{19.53} & \textbf{22.71} & \textbf{20.02} & \textbf{19.38} \\ \midrule
\multirow{2}{*}{Flash} & \checkmark &  &  &  & 33.13 & 24.32 & 20.49 & 30.97 & 21.74 & 17.74 \\
 & \checkmark & \checkmark & \checkmark & \checkmark & \textbf{48.78} & \textbf{32.98} & \textbf{26.82} & \textbf{45.43} & \textbf{32.59} & \textbf{26.28} \\ \bottomrule
\end{tabular}}
    \caption{Comparison of the CMA model using different combination of cross-embodiment training data.}
    \label{tab:cross_data_training}
    \end{table}

\subsection{Impact of Lighting Conditions}
\begin{figure}[tb]
    \centering
    \includegraphics[width=0.95\linewidth]{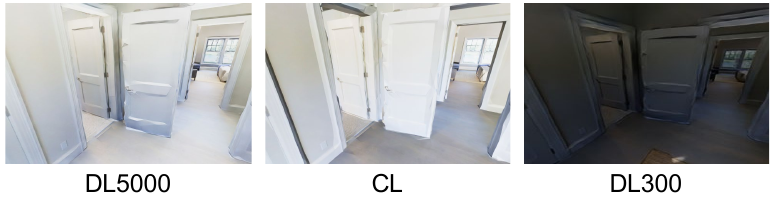}
    \caption{Environments under different light conditions.}
    \label{fig:light}
\end{figure}
Omniverse Kit-based applications provide versatile lighting options to simulate various environments. We use three settings: \texttt{Disc Light} (DL) – simulating panel lighting, with intensity set to 5000 for daytime and 300 for nighttime; and \texttt{Camera Light} (CL) – simulating light from the robot. This may cause reflections on the front wall, leaving the sides darker.
As shown in Tab.~\ref{tab:light_condition}, NaVid suffers the largest performance drop when lighting deviates from ideal conditions, with SR decreasing by 12.47\% under DL300 and 11.25\% under CL on val-unseen.
In contrast, CMA and RDP are less affected. The key reason behind this discrepancy lies in modality reliance: NaVid relies solely on RGB input, making it highly susceptible to lighting variations, whereas CMA and RDP utilize both RGB and depth, allowing them to maintain stability when RGB quality degrades.
This underscores the need to integrate multimodal environmental information, such as depth or radar, to improve model generalization and robustness. 
\begin{table}[t]
\centering
\renewcommand{\arraystretch}{0.85}
\resizebox{\linewidth}{!}{
\begin{tabular}{@{}l|l|ccc|ccc@{}}
\toprule
\multicolumn{1}{l|}{\multirow{2}{*}{Method}} & \multirow{2}{*}{Light} & \multicolumn{3}{c|}{Val   Seen} & \multicolumn{3}{c}{Val Unseen} \\
\multicolumn{1}{l|}{} &  & \multicolumn{1}{c}{OS$\uparrow$} & \multicolumn{1}{c}{SR$\uparrow$} & \multicolumn{1}{c|}{SPL$\uparrow$} & \multicolumn{1}{c}{OS$\uparrow$} & \multicolumn{1}{c}{SR$\uparrow$} & \multicolumn{1}{c}{SPL$\uparrow$} \\ \midrule
\multirow{3}{*}{NaVid (ZS)} & DL5000 & 24.32 & 21.58 & 17.45 & 27.32 & 22.42 & 18.58 \\
 & DL300 & 17.02 & 14.59 & 12.19 & 12.47 & 9.95 & 9.01 \\
 & CL & 21.23 & 18.38 & 16.10 & 14.27 & 11.17 & 9.34 \\ \midrule
\multirow{3}{*}{CMA} & DL5000 & 31.04 & 21.12 & 16.15 & 31.33 & 18.78 & 14.56 \\
 & DL300 & 30.06 & 19.02 & 15.55 & 30.49 & 17.37 & 15.34 \\
 & CL & 28.23 & 19.13 & 14.45 & 29.02 & 17.04 & 15.28 \\ \midrule
\multirow{3}{*}{RDP} & DL5000 & 33.43 & 23.86 & 17.35 & 34.06 & 21.98 & 16.44 \\
 & DL300 & 29.94 & 20.52 & 15.71 & 29.40 & 22.27 & 17.15 \\
 & CL & 29.87 & 21.06 & 16.89 & 30.87 & 22.78 & 16.23 \\ \bottomrule
\end{tabular}}
\caption{Comparison of using different light conditions.}
\label{tab:light_condition}
\end{table}

\section{Conclusion}
\label{sec: conclusion}
We introduce VLN-PE, a realistic VLN platform and benchmark designed to enhance physical deployment across diverse robot embodiments. It enables cross-embodiment data collection, evaluation, and optimization under realistic locomotion and environmental conditions. Through systematic experiments on VLN methods based on the ego-centric pinhole cameras, we expose critical physical and visual disparities that challenge existing approaches and benchmarks. VLN-PE offers a grounded framework to foster more generalizable VLN models for future physical embodied AI development.

\section*{Acknowledgments}
We sincerely thank all our collaborators for their valuable support and contributions. This paper is supported by the National Natural Science Foundation of China under Grants (624B2105, 62473295, 62233013, 62333017). This work is funded in part by the National Key R\&D Program of China (2022ZD0160201), and Shanghai Artificial Intelligence Laboratory.
{
    \small
    \bibliographystyle{ieeenat_fullname}
    \bibliography{main}
}
\clearpage \appendix
\section*{Appendix}
\section{Implementation Details}
Since the most commonly used mobile robots are equipped with ego-centric RGB-D pinhole cameras currently, we primarily evaluate VLN methods without panoramic views in this work. For classic end-to-end single-step discrete action prediction models (Seq2Seq~\cite{krantz_beyond_2020}, CMA~\cite{krantz_beyond_2020}, and NaVid~\cite{zhang2024navid}), we directly use their publicly available code and pre-trained weights. For the other two model types—the end-to-end continuous multi-step prediction model (RDP) and the map-based LLM model (VLMaps~\cite{huang23vlmaps})—we introduce several modifications, which are detailed in this appendix. 

\subsection{Recurrent Diffusion Policy for VLN}
\label{subsec_vln_dp}
The RDP model takes ego-centric RGB-D observations and language instructions as inputs. Since some instructions exceed the 77-word limit in standard CLIP~\cite{radford2021learning}, LongCLIP~\cite{zhang2024long} is used as both the RGB and instruction encoders. The depth encoder follows CMA, using ResNet50 pre-trained on point-goal tasks. Each RGB image is represented by five tokens: The first token encodes the global feature, while the remaining four tokens capture semantic information via grid pooling~\cite{zhang2024navid}. The flattened depth features are added to the first RGB token, resulting in a fused visual feature dimension of $\mathbb{R}^{5\times h_d}$, where $h_d=512$.
To improve progress awareness, we incorporate previous 4-step actions ($PA$) and relative coordinates ($RC$) from the starting point, both represented in \((\Delta x, \Delta y, \Delta yaw)\). The key difference is that $PA$ encodes the last four steps relative to the current position, while $RC$ represents the current position relative to the starting point.

For historical observation encoding, initially, we experimented with a video-based format, similar to NaVid, where stacked images provided long-term sequence information. However, this approach led to rapid convergence of the diffusion process to small losses, causing severe overfitting. Through our experimentation, we found that employing a recurrent GRU structure to maintain and update historical observations improved generalization:
\begin{align}
    h_t=\text{GRU}([V_c, RC,PA], h_{t-1}).
\end{align}

Then, we apply two cross-attention mechanisms to align attended vision ($q=\text{Concat}(h_t,V_c)$) and language features $I=\{w_i\}_{i=1}^{L}$, where each modality serves as the key and value for the other:
\begin{align}
    g_1&=\text{CrossAttn}(q,I,I),\quad g_2=\text{CrossAttn}(I,q,q).
\end{align}

\begin{figure}
    \centering
    \includegraphics[width=\linewidth]{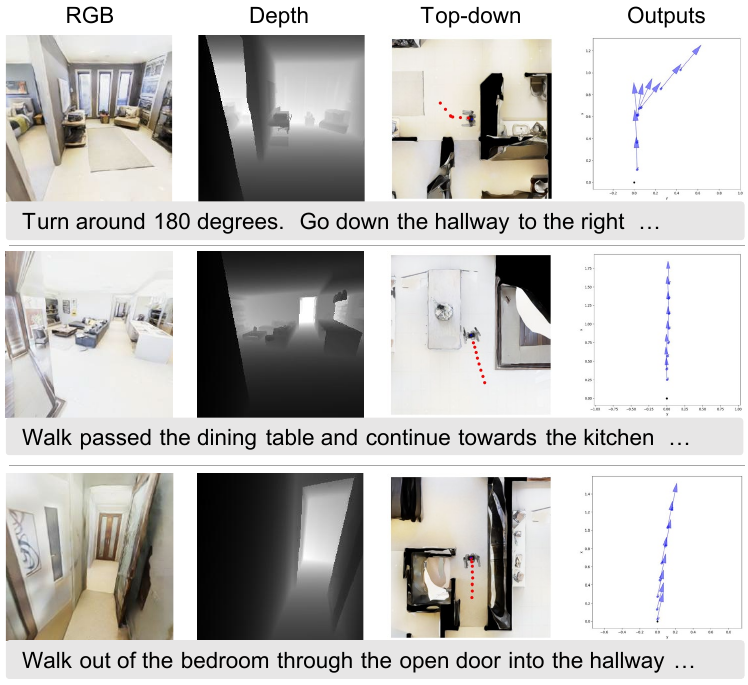}
    \caption{Examples of the robot observations and RDP outputs.}
    \label{fig:rdp_outputs}
\end{figure}

Finally, the condition feature for the diffusion model is formed by concatenating all extracted features: 
\begin{align}
    c_t=\text{Concat}(g_1,g_2,h_t,RC,PA).
\end{align}
We employ a transformer-based diffusion module~\cite{sridhar2024nomad} with one encoder layer and three decoder layers. During training, the ground-truth trajectory coordinates $T\times (\Delta x, \Delta y, \Delta yaw)$ are perturbed with random noise, and the network is trained to predict and remove this noise. The iterative denoising process follows DDPM~\cite{ho2020denoising}. 
Additionally, we introduce a self-attention-based stop prediction head to determine the current stop progress (from $0$ to $1$). The stop signal is triggered if: All predicted actions from the diffusion head are below the threshold 0.1, or the stop progress output exceeds 0.8. The output of the RDP is shown in Fig.~\ref{fig:rdp_outputs}. During navigation, RDP predicts 8 future trajectory waypoints and executes 4 steps per iteration.

In our experiments, RDP demonstrated improvements over the previous baseline models (Seq2Seq and CMA) when trained from scratch. However, there remains significant potential for further enhancement. As this paper primarily focuses on the new physical VLN platform (VLN-PE), we introduce RDP as a baseline method for predicting trajectory waypoints, which can be further integrated with control-theoretic approaches like the Model Predictive Control (MPC) framework to enhance motion smoothness, addressing the jerky transitions seen in discrete action-based methods. We hope this work can inspire and support some future research in this direction.

\subsection{Improved VLMaps}
\label{subsec_VLMaps}
\begin{figure}
    \centering
    \includegraphics[width=\linewidth]{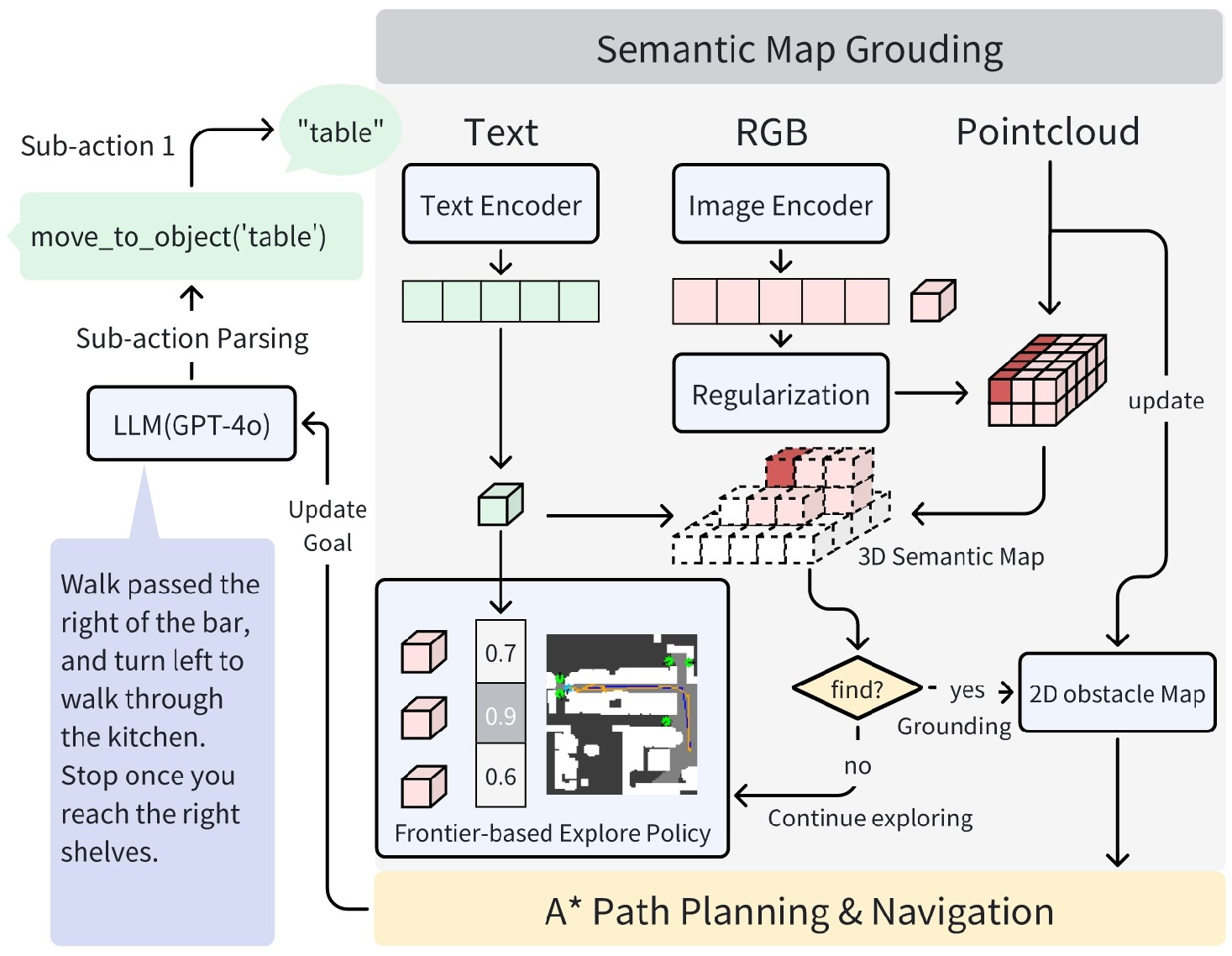}
    \caption{Framework of the improved VLMaps.}
    \label{fig:vlmaps}
\end{figure}
VLMaps differs from traditional end-to-end models by utilizing a spatial semantic map representation that directly integrates pre-trained vision-language features of the physical world. This approach enables natural language-based map indexing without requiring additional labeled data. Therefore, we chose this method as one of the technical pipelines for evaluation. However, the original VLMaps lacks a direct exploration policy and struggles with room-level descriptions (\textit{e.g.}, ``enter the living room''), which require an agent-oriented perspective rather than reliance on a global semantic map. To address these limitations, we improved the VLMaps framework (as shown in Fig.~\ref{fig:vlmaps}) with two key enhancements.

\textbf{Exploration Policy:} Inspired by VLFM~\cite{yokoyama2024vlfm}, we implement a frontier detection strategy, where a frontier is defined as the boundary between explored and unexplored areas. When the robot fails to detect the target object from the current pixel embeddings, it performs a turn-around maneuver, moves to each frontier, and evaluates whether the viewpoint is likely to contain the next landmark using the image and text encoders from LSeg. For instance, we observed that ``table'' exhibits a higher similarity score with scenes of ``dining room'' compared to ``toilet,'' validating the policy's ability to guide the robot toward plausible directions.

\textbf{Room-Level Descriptions:} Similarly, we leverage the CLIP module from LSeg as a classifier to assess whether the viewpoint aligns with the intended room context. Specifically, we use a predefined set of room names (``living room,'' ``dining room,'' ``bedroom,'' ``kitchen,'' ``toilet,'' ``others'') as text inputs to index the current RGB image. Upon successful room detection, we naturally incorporate actions such as \texttt{self.move\_to\_room('room\_name')}.

\begin{figure*}
    \centering
    \includegraphics[width=\linewidth]{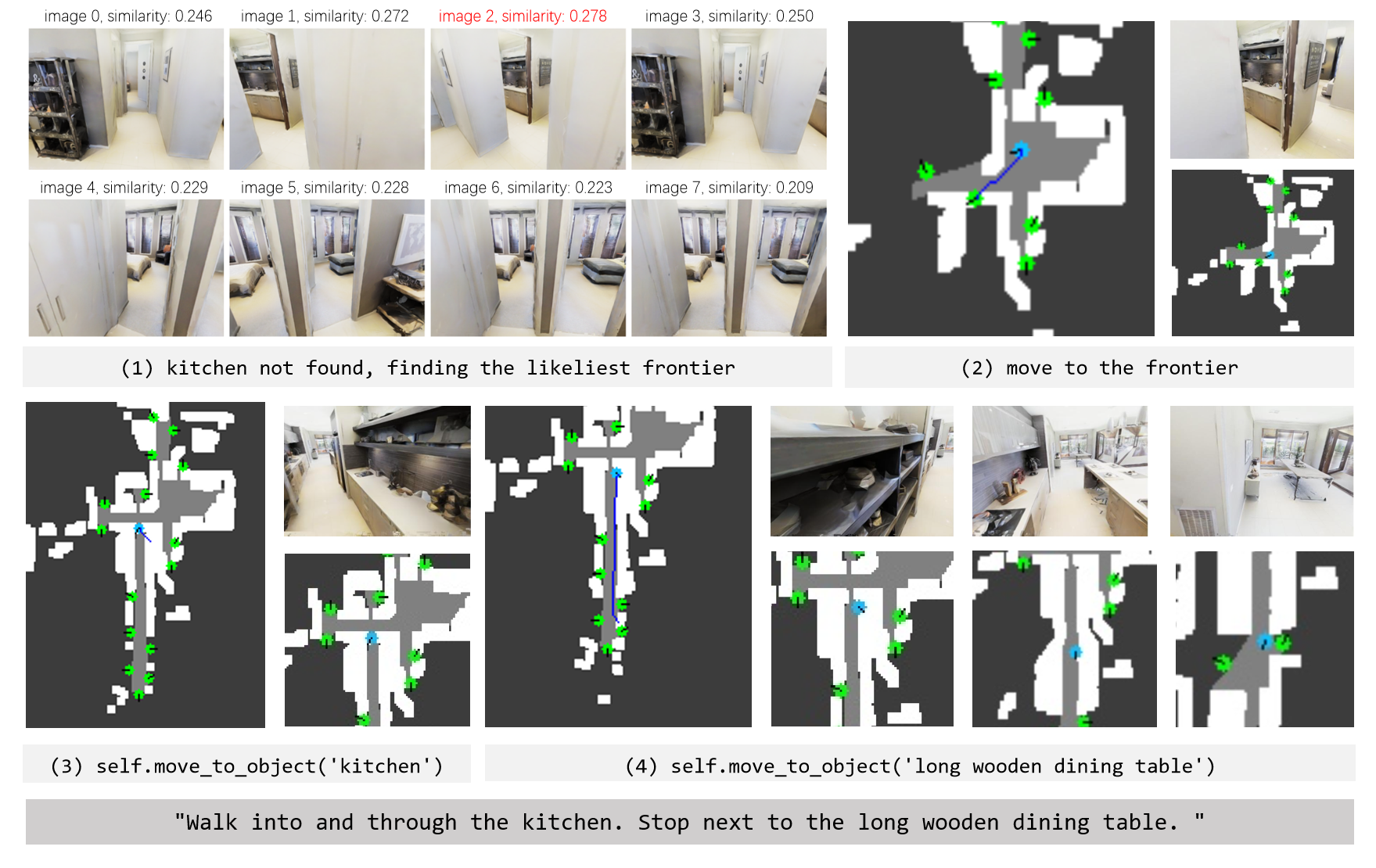}
    \caption{Example of improved VLMaps. Blue dot: current position (black line: orientation). Green dot: frontiers (black line: exploration orientation). White: dilated obstacles. Light gray: explored area. Dark gray: unexplored area. Blue line: local planner trajectory.}
    \label{fig:vlmaps_outputs}
\end{figure*}

In the VLN-PE, we apply additional techniques to reduce the fall rate and stuck rate. For example, we implement an A* algorithm as the local planner, assigning higher costs to dilated and unexplored areas. When executing commands like \texttt{self.move\_forward(1)}, the robot may collide with obstacles if not properly oriented. To address, we define a cost function to identify the optimal node $n^*$ from the robot's perspective: $n^*=\mathop{\arg\min}\limits_{n}(||dist(n,x_0)-dist(x_g,x_0)||+\alpha\gamma)$, where \( x_g \) is the goal position, \( x_0 \) is the current position, \( \alpha =0.25\) is a weight parameter, and \( \gamma \) is the angle required to face the target. This ensures the robot slightly reorients itself before moving forward, minimizing collision risks. An example of the improved VLMaps is shown in Fig.~\ref{fig:vlmaps_outputs}. Compared to end-to-end methods, map-based modular approaches offer more explainable and reliable results. However, their performance heavily depends on mapping and localization accuracy, which could limit the practical deployment.

\subsection{Experimental Details}
All training experiments are conducted using NVIDIA RTX 4090 GPUs. The CMA and Seq2Seq models are trained on a single GPU with a batch size of 2, requiring approximately one day to converge. The RDP model is trained on 4 GPUs using PyTorch's DataParallel module, with a total batch size of 8, and completes training in around two days. All models are optimized using the AdamW optimizer with a learning rate of $1 \times 10^{-4}$. The maximum trajectory length is set to 200. For evaluation, the CMA model requires approximately 4 hours to complete a full evaluation on the R2R-CE benchmark when run in parallel on 8 GPUs.

\subsection{Datasets}
\begin{figure}[htbp]
    \centering
    \begin{minipage}[t]{0.49\linewidth}
        \centering
        \includegraphics[width=\linewidth]{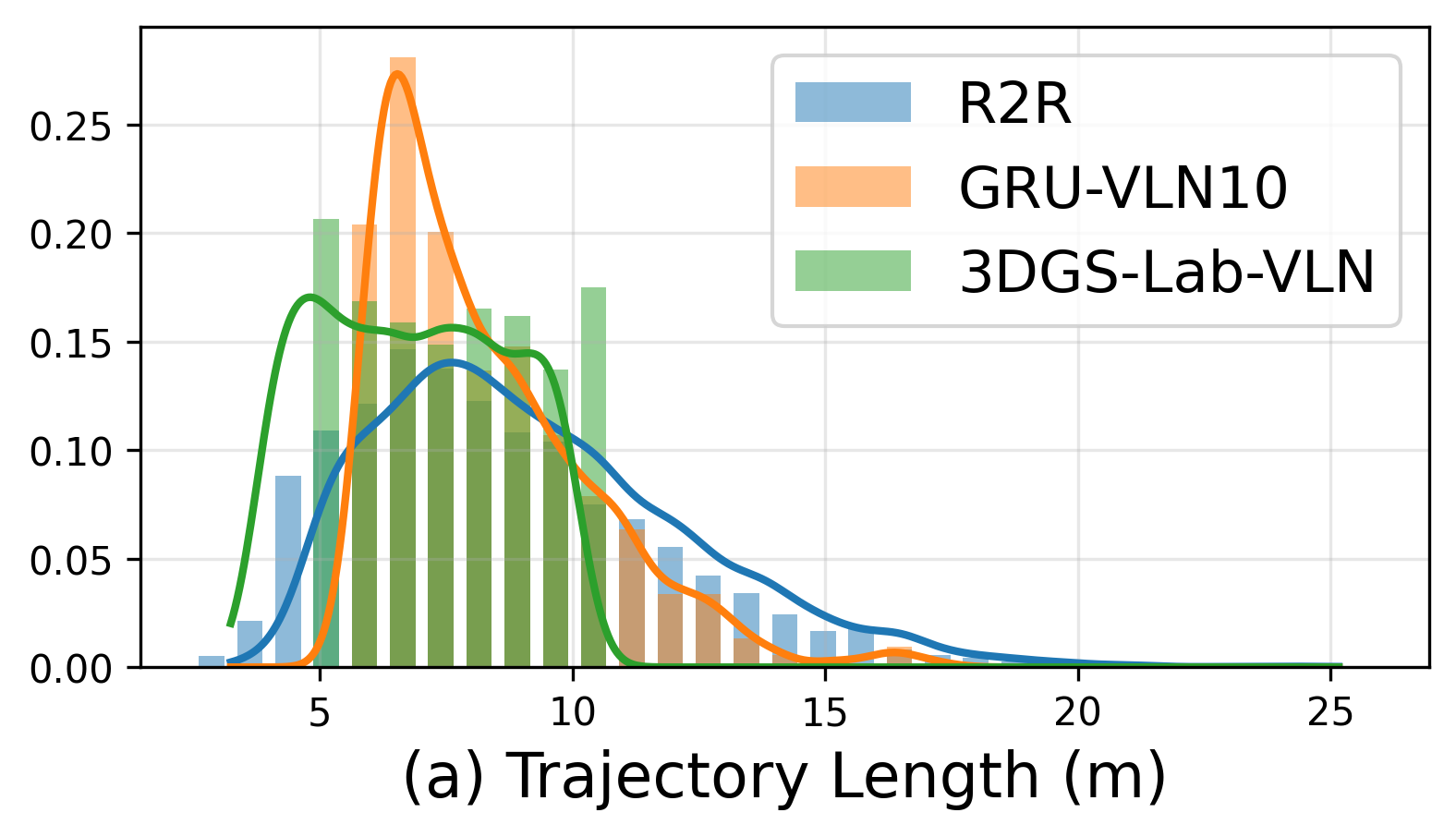}
        \label{fig_traj_length}
    \end{minipage}
    \hfill
    \begin{minipage}[t]{0.49\linewidth}
        \centering
        \includegraphics[width=\linewidth]{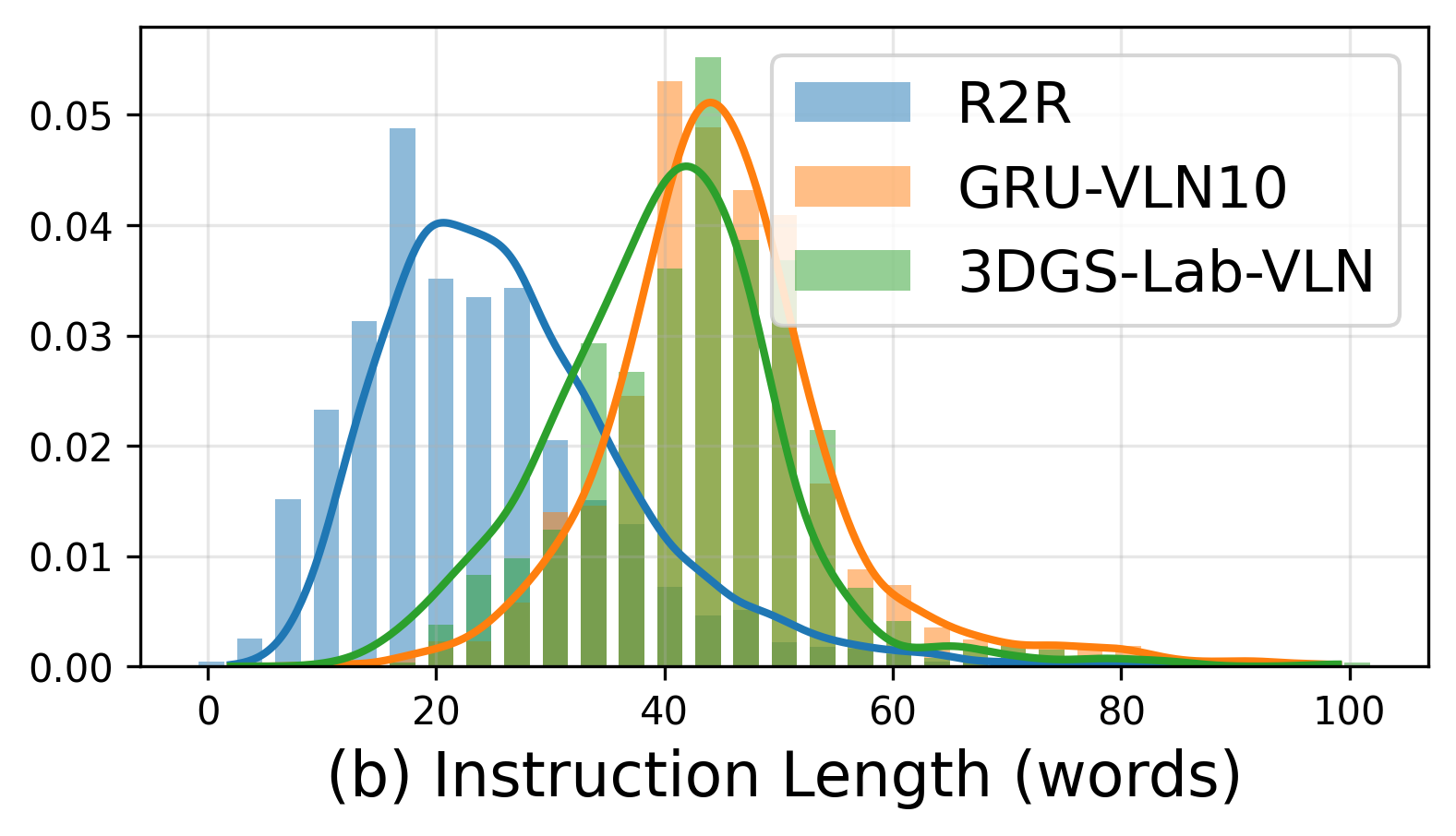}
        \label{fig_instr_length}
    \end{minipage}

    \vspace{-1em} 

    \begin{minipage}[t]{0.32\linewidth}
        \centering
        \includegraphics[width=\linewidth]{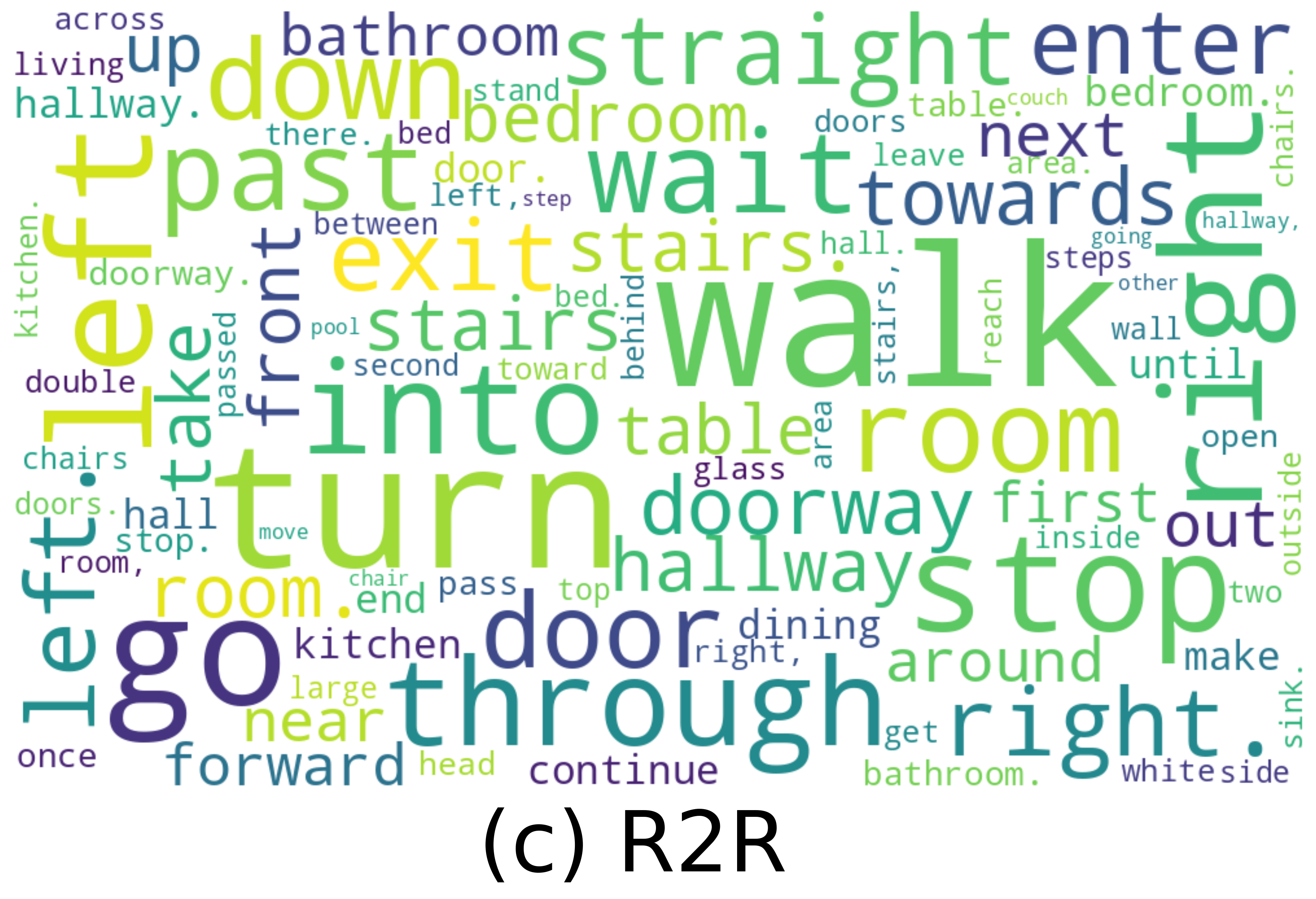}
        \label{fig_word_cloud_r2r}
    \end{minipage}
    \hfill
    \begin{minipage}[t]{0.32\linewidth}
        \centering
        \includegraphics[width=\linewidth]{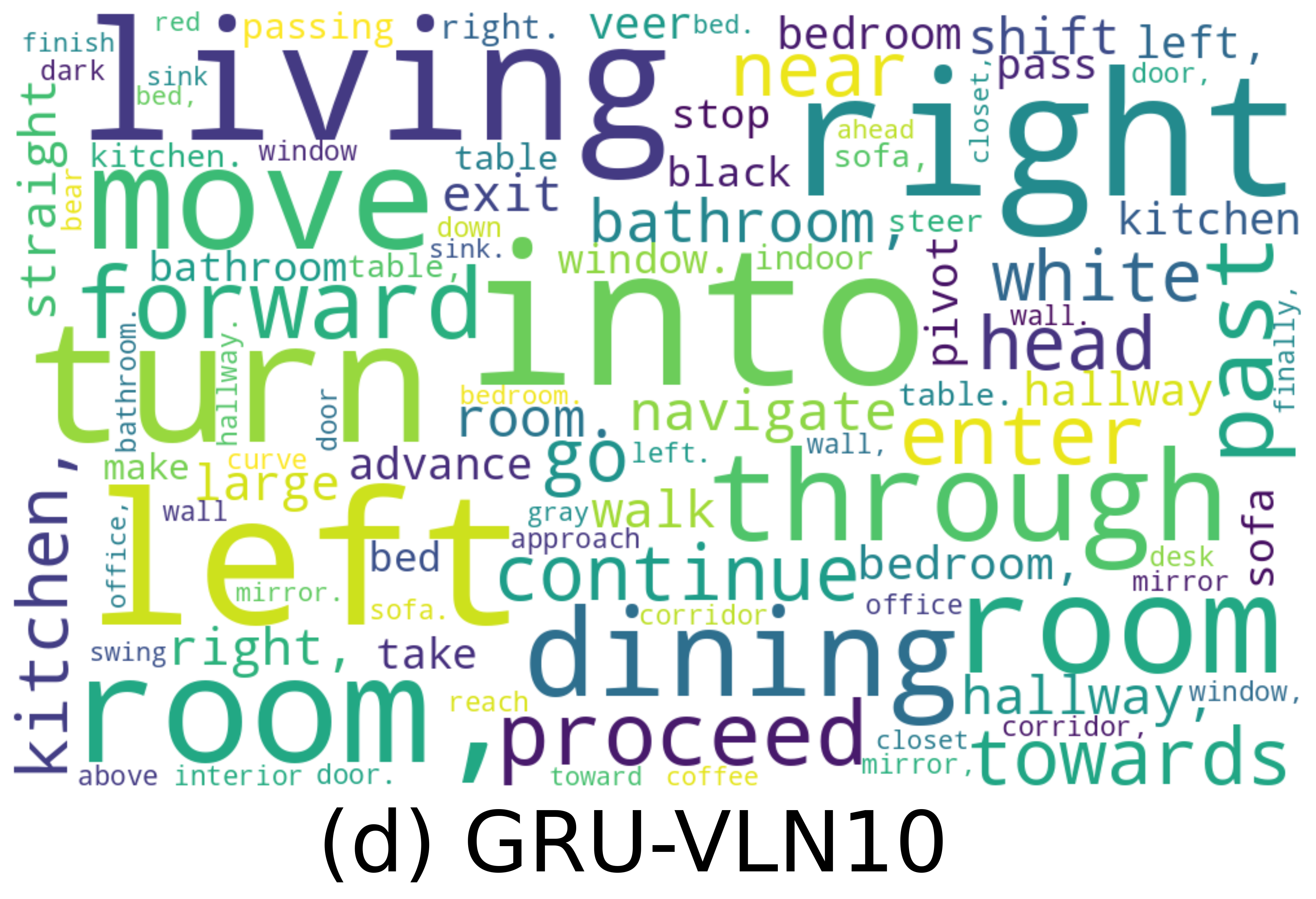}
        \label{fig_word_cloud_gruvln10}
    \end{minipage}
    \hfill
    \begin{minipage}[t]{0.32\linewidth}
        \centering
        \includegraphics[width=\linewidth]{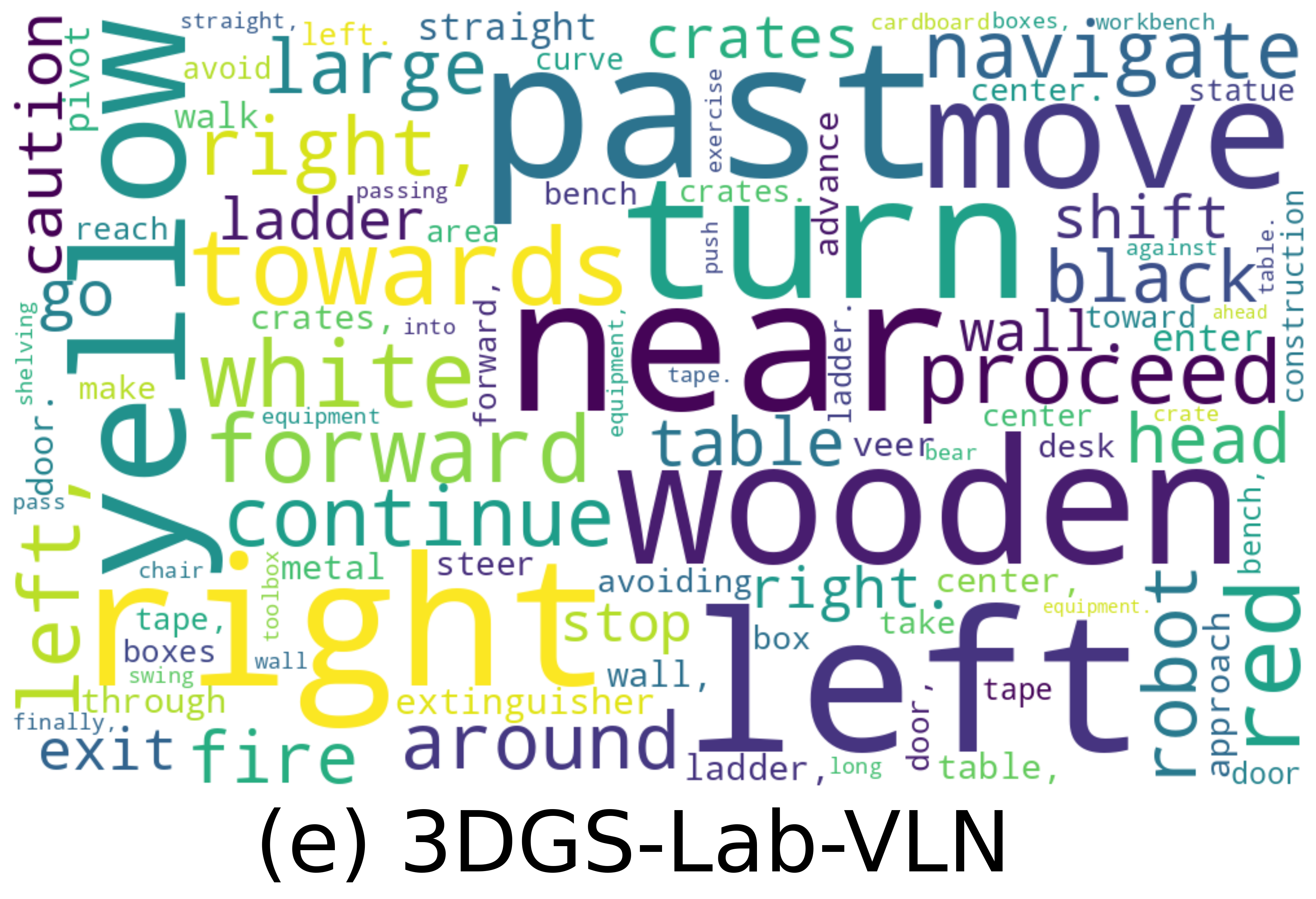}
        \label{fig_word_cloud_3dgs}
    \end{minipage}
    \vspace{-1em}
    \caption{Comparison of distributions across datasets.}
    \label{fig_combined_stats}
\end{figure}

The trajectories sampling strategy for the newly introduced datasets (GRU-VLN10 and 3DGS-Lab-VLN) is as follows: (a) generate a freemap, (b) randomly sample start-goal pairs, and (c) filter out invalid paths (overly short, long, or similar ones). Instructions are generated via a modular pipeline~\cite{he2025navcomposercomposinglanguageinstructions} with action and environment recognition, GPT-4 in-context description, and human refinement. Comparisons of datasets are presented in Fig.~\ref{fig_combined_stats}.

\subsection{Metrics}
{\bf Metrics.} Following standard VLN evaluation protocols~\cite{anderson2018vision, krantz_beyond_2020}, we use five primary metrics: \textit{Trajectory Length (TL)}, measured in meters; \textit{Navigation Error (NE)}, which quantifies the distance between the predicted and actual stop locations; \textit{Success Rate (SR)}, indicating how often the predicted stop location falls within a predefined distance of the true location; \textit{Oracle Success Rate (OS)}, which assesses the frequency with which any point along the predicted path is within a certain distance of the goal; and \textit{Success Rate weighted by Inverse Path Length (SPL)}, which balances success rate with path efficiency. 
As physical realism is a key focus of this work, we introduce two more metrics: \textit{Fall Rate (FR)}, which measures the frequency of unintended falls, and \textit{Stuck Rate (StR)}, which quantifies instances where the agent becomes immobilized. Specifically, \textit{``Fall"} is the robot having a roll $>$ 15° or pitch $>$ 35°, or a center-of-mass–to-foot height below a robot-specific threshold. \textit{``Stuck"} is defined as both position and heading change $<$ 0.2m and 15° for 50 steps.

\subsection{Controllers}
\label{subsec_appendix_controller}
Thanks to NVIDIA Isaac Sim's advanced physical simulation capabilities, we can seamlessly apply various control theories, making the low-level control policy more diverse and aligned with real-world robotic applications. In this work, we utilize three types of controllers for experimentation: flash control, move-by-speed control, and move-along-path control.

\begin{itemize}
\item Flash Control: This mechanism mimics platforms that lack physical cross-embodiment support, allowing the agent to instantly reach the target position without considering physical motion constraints.
\item Move-by-Speed Control: This method simulates realistic motion dynamics by controlling the agent’s velocity using linear and angular speed commands. For legged humanoid and quadruped robots, we employ the RL-based policies to regulate movement, ensuring the robot follows the required forward and rotational speeds. For wheeled robots, we use a differential drive controller to manage navigation. For end-to-end models, we implement discrete actions using this controller.
\item Move-along-Path Control: This approach enables the agent to follow a predefined trajectory, replicating path-following behaviors in robotic navigation. For the Map-based method (VLMaps), we apply the A* path planning algorithm and use this controller with a PID system to ensure smooth trajectory following.
\end{itemize}

\subsection{Fine-tune on the specific datasets}
To better evaluate out-of-MP3D-style domain generalization, we collect additional VLN datasets using GRUScenes and 3DGS-rendered environments. Since these datasets are primarily used for evaluation, only a small portion of the data is allocated for training, while the majority is reserved for testing. For CMA and RDP, all training experiments use a learning rate of 1e-4 with a cosine learning schedule. In Tab.~\ref{tab:gru_vln10} and Tab.~\ref{tab:3dgs_performance}, ``w/o FT" refers to direct zero-shot transfer using VLN-PE-R2R-trained weights for evaluation without further in-domain fine-tuning in these new scenes. On the GRU-VLN10 dataset, small models significantly improved after 10 epochs of fine-tuning, whereas the SoTA large model NaVid showed limited zero-shot performance. This highlights the limited diversity of existing VLN benchmarks, which could not fully assess model generalization. 

\section{Impact of Sim-to-Real Transfer}
\begin{figure}[bp]
    \centering
    \includegraphics[width=\linewidth]{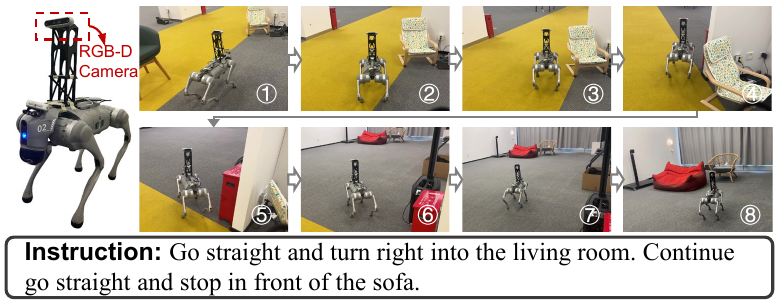}
    \caption{Real-world experiments using a Unitree Go2 robot.}
    \label{fig_real_test}
\end{figure}

Compared to traditional VLN simulators and platforms, VLN-PE introduces a significant advancement by supporting physical VLN across diverse robot types, enabling data collection, training, and closed-loop evaluation in physical settings. We begin by identifying the limitations of existing VLN algorithms when deployed in physical environments, as initially verified within a physics-enabled simulator. After fine-tuning on data collected through VLN-PE, we observe consistent performance improvements within the simulated physical setup.
Encouraged by these results, we further evaluate our approach in real-world settings. Specifically, we conduct experiments using a Unitree Go2 robot equipped with an Intel RealSense D455 RGB-D camera across 14 indoor episodes (see Table~\ref{tab_real_test} and Fig.~\ref{fig_real_test}). The model fine-tuned with VLN-PE demonstrates improved adaptation and generalization, confirming the practical effectiveness of our platform.
\renewcommand{\arraystretch}{0.6}
\begin{table}[bp]
\centering
\footnotesize
\begin{tabular}{@{}c|c|cc@{}}
\toprule
Method               & Fine-tuned on VLN-PE  & OS↑   & SR↑   \\ \midrule
\multirow{2}{*}{CMA} & \ding{55}                     & 14.29 & 7.14  \\
                     & \ding{51}                     & \textbf{57.14} & \textbf{28.57} \\ \bottomrule
\end{tabular}
\caption{Impact of VLN-PE on real-world performance.}
\label{tab_real_test}
\end{table}

In particular, we observe that the CMA baseline model, after VLN-PE fine-tuning, exhibits more confident forward movement and better semantic grounding during navigation. In contrast, the CMA Full baseline trained solely on VLN-CE struggles in real-world conditions, frequently resulting in aimless rotation and poor generalization. One notable remaining challenge is the handling of the stop action. CMA often fails to robustly predict when to stop. To mitigate this, we let the robot output the stop action when the predicted probability of the stop action exceeds $1 \times 10^{-4}$.

\section{Analysis of Failure Cases}
\begin{figure}[htbp]
    \centering
    \includegraphics[width=\linewidth]{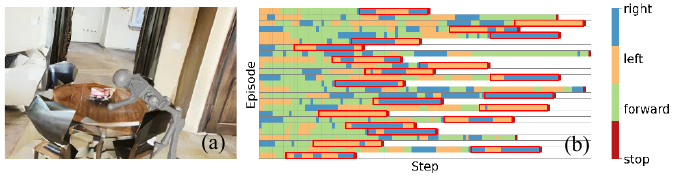}
    \caption{Visualization of the failure cases. (a) shows a typical failure case where the agent collides and falls. (b) highlights NaVid's repetitive turning before stopping (red box).}
    \label{fig_failure_case}
\end{figure}

\begin{figure*}
    \centering
    \includegraphics[width=\linewidth]{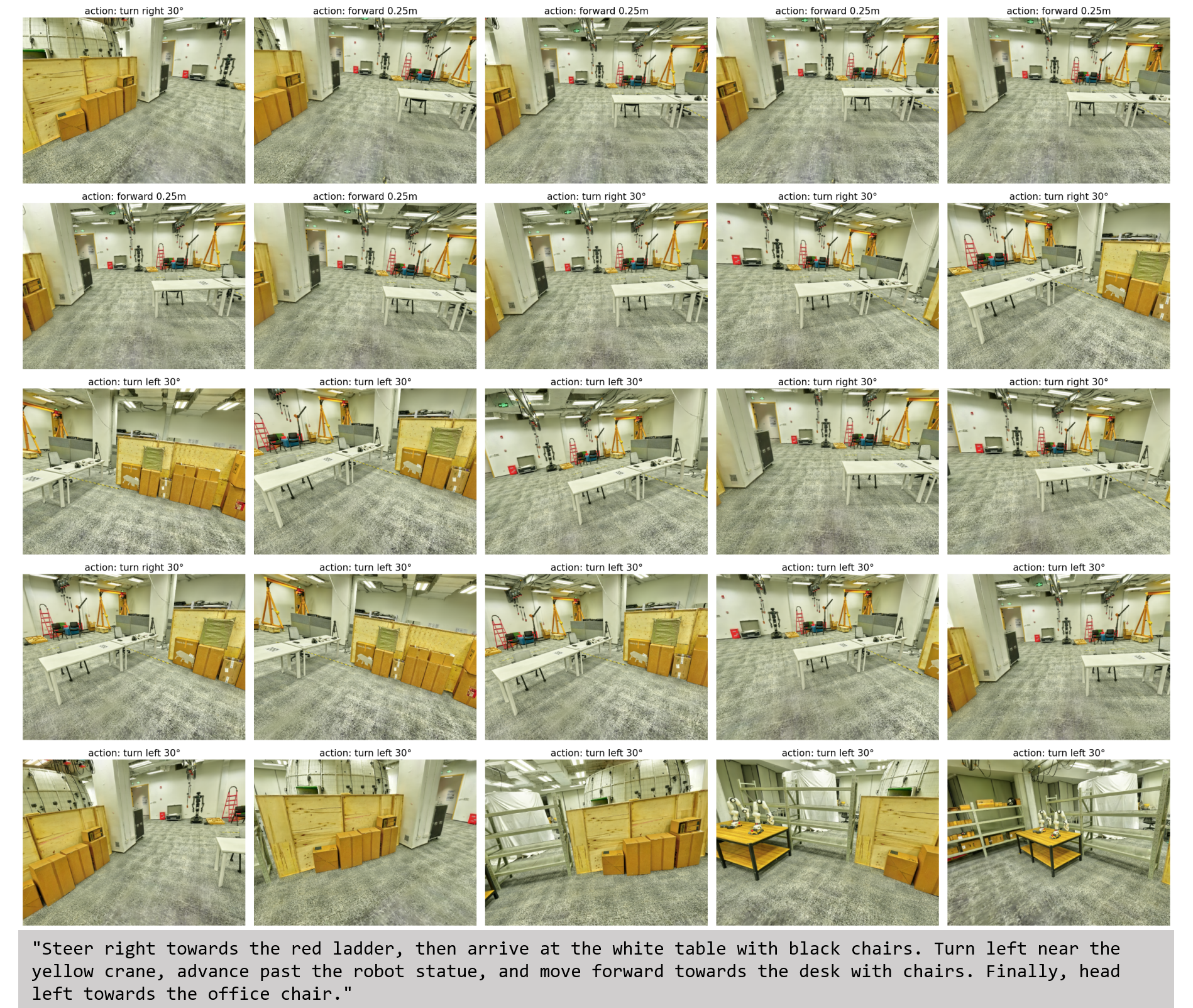}
    \caption{Visualization of the failure case for the ego-centric SoTA VLN model, NaViD, in 3DGS online-rendered scenes. The model tends to predict rotation actions, indicating a failure to interpret the intended trajectory.}
    \label{fig:navid_anaylze}
\end{figure*}

In Tab.~\ref{tab:3dgs_performance}, we observe that the SoTA ego-centric model, NaViD \cite{zhang2024navid}, shows exceptionally poor zero-shot performance (\textit{e.g.,} 5.8 SR and 1.0 SPL) on our 3DGS-Lab-VLN datasets. The possible reasons for the performance degradation could be summarized as follows: Firstly, the use of 3D Gaussian Splatting (3DGS) for rendering may introduce artifacts and distortions. As shown in Fig.~\ref{fig:navid_anaylze}, rendering artifacts can cause some blurring in ground areas and distant details, introducing subtle distortions that may go unnoticed by the human eye. In our experiments, the model relying solely on RGB input is highly vulnerable to such pixel-level noise, leading to failure in affected scenes. This underscores the need for research on image perturbations and related safety issues in VLN models.
Additionally, we note that the NaViD model frequently rotates in circles to find a better viewpoint for localizing the target, which accounts for 70\% of failures (Fig.~\ref{fig_failure_case}).
In summary, our findings on the limitations of current SoTA VLN methods align with the conclusions of this paper. We hope our insights and tools will drive the development of more robust and generalizable VLN models, especially in diverse, non-MP3D-style environments.

\section{Limitations and Future Work}
While this work evaluates various ego-centric VLN methods, some other state-of-the-art VLN approaches rely on panoramic observations~\cite{an2024etpnav,anderson2021sim,krantz2022sim}. These methods use panoramic views with depth to generate sparse waypoint connections, integrating them with discrete VLN techniques for path selection—an approach that has demonstrated strong performance in previous non-physical settings. 
Since our primary goal is to evaluate the existing VLN methods under the physical settings, we adopt an ego-centric view setting to align with current robotic perception systems. However, as robotics evolves—potentially resembling autonomous driving systems with more diverse RGB or radar sensors—future robots may benefit from panoramic perception. Thus, we plan to extend our evaluation to panoramic VLN methods in future work. Additionally, with multi-robot support and real-time 3DGS scene rendering, our platform has significant potential to facilitate a real-sim-real VLN pipeline, enhancing real-world adaptability for embodied agents in familiar environments. We leave this for future research.

\section{Additional Qualitative Examples}
\label{subsec_visualization}
\begin{figure*}
    \centering
    \includegraphics[width=\linewidth]{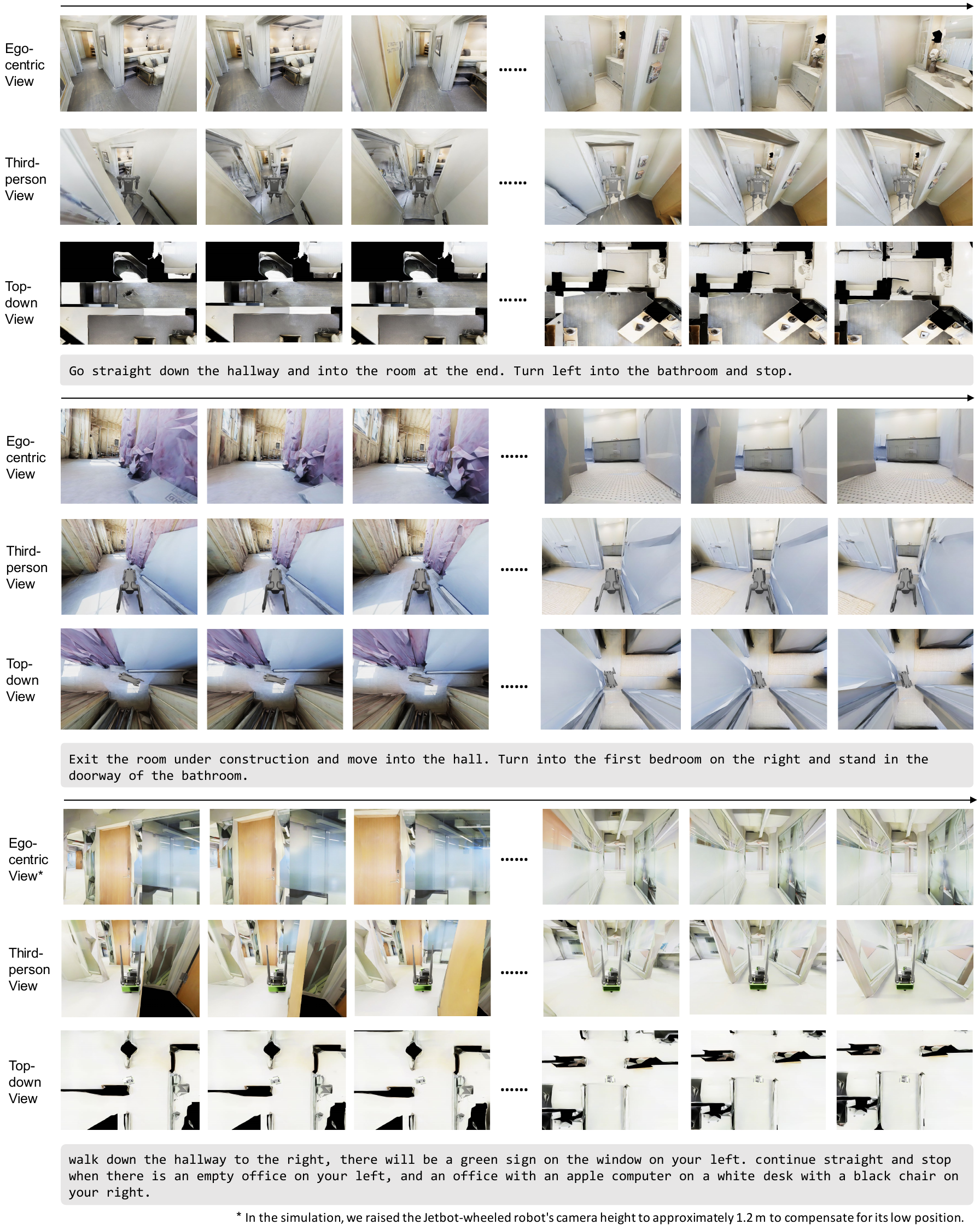}
    \caption{Visualization of different robot viewpoints in VLN-PE. Leveraging the powerful interactive capabilities of Isaac Sim, researchers can easily observe robot motion from various perspectives within the environment.}
    \label{fig:appendix_robot_views}
\end{figure*}

\begin{figure*}
    \centering
    \includegraphics[width=\linewidth]{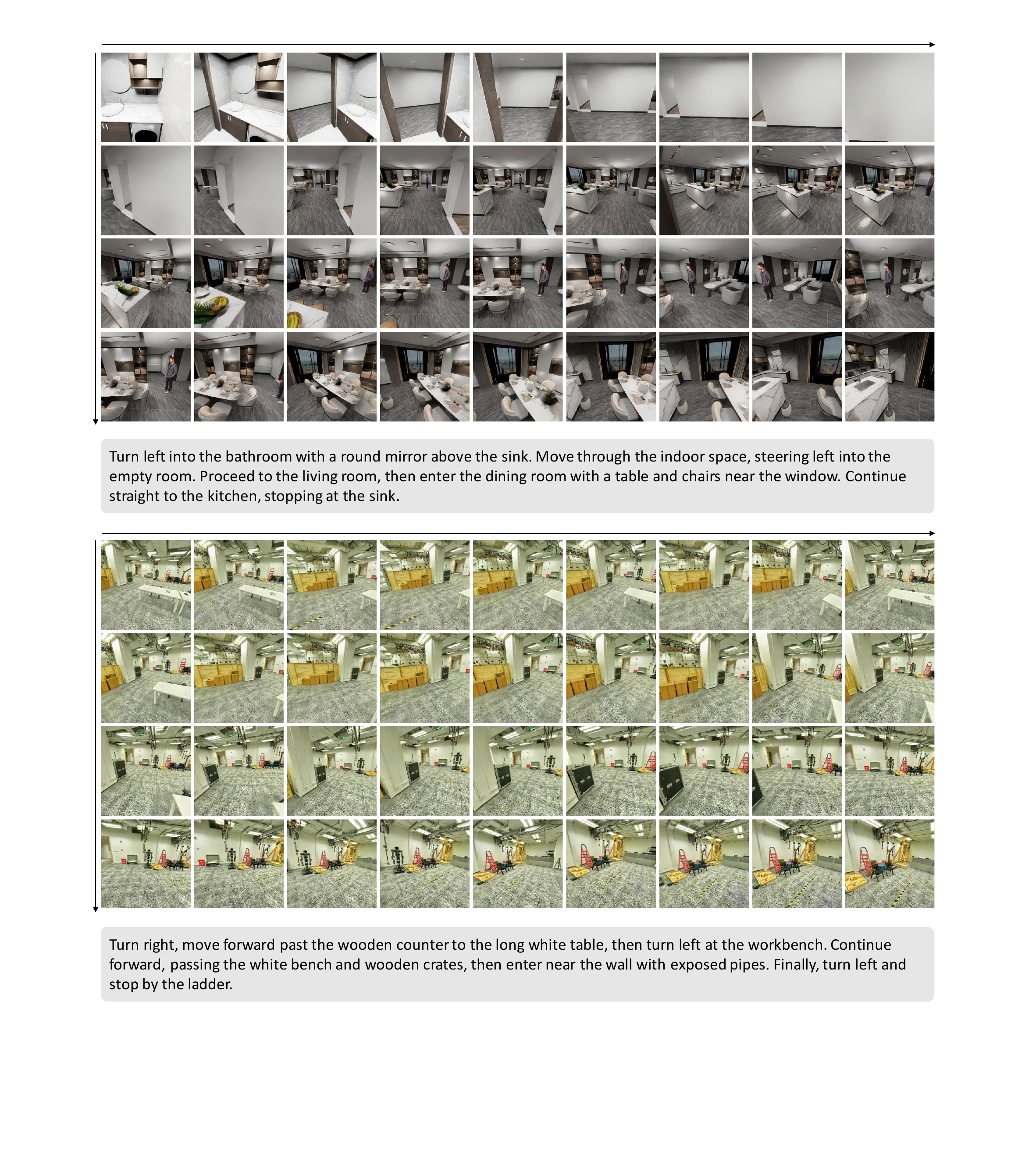}
    \caption{Examples of trajectories and instructions from our introduced GRU-VLN10 and 3DGS-Lab-VLN datasets.}
    \label{fig:appendix_gruvln10_sixth_floor}
\end{figure*}

To better illustrate the observations and environments within VLN-PE, we provide supplementary videos showcasing the significant shaking and instability experienced by physical agents during navigation. Additionally, Fig.~\ref{fig:appendix_robot_views} presents different viewpoints—ego-centric, third-person, and top-down—using various robot types in VLN-PE. Fig.~\ref{fig:appendix_gruvln10_sixth_floor} displays trajectories and instructions from our newly introduced 10 high-quality synthetic scenes (GRU-VLN10) and a 3DGS online-rendered scene (3DGS-Lab-VLN), supporting out-of-MP3D-style evaluations.

\end{document}